\documentclass[letterpaper, 10pt, conference]{ieeeconf}  

\IEEEoverridecommandlockouts                              

\usepackage{amsmath} 
\usepackage{amssymb}  
\makeatletter
\let\NAT@parse\undefined
\makeatother
\usepackage[numbers]{natbib}
\let\labelindent\relax
\usepackage{enumitem}

\usepackage{url}
\usepackage{amsmath} 

\usepackage{subcaption}
\usepackage{tikz}
\usepackage{pgfplots}
\usepackage{pgfplotstable}
\usepgfplotslibrary{groupplots}

\pgfplotsset{
   enlarge x limits=0,
   enlarge y limits=0,
   tick label style={/pgf/number format/fixed, font=\tiny}, 
   log ticks with fixed point,
   ylabel style={yshift=-0.65cm, font=\footnotesize, align=center},
   xlabel style={yshift=0.25cm, font=\footnotesize},
   title style={yshift=-0.15cm, font=\footnotesize},
   xticklabel style={yshift=0.05cm, font=\scriptsize},
   yticklabel style={xshift=0.05cm},
   grid style={line width=.1pt, draw=gray!10},
   major grid style={line width=.2pt,draw=gray!40},
   legend style={draw=none, 
    			  inner sep=0pt, outer sep=0pt,
				  every node/.style={inner sep=0pt, outer sep=0pt},
        	     /tikz/every even column/.append style={column sep=0.75em},  
        		  font=\footnotesize, anchor=north}
}

\usepackage{booktabs}
\usepackage{tabularx}
\usepackage{makecell}
\usepackage{multirow}
\usepackage[bookmarks=true]{hyperref}

\newcommand{\Chi}{\mathrm{X}}
\newcolumntype{Y}{>{\centering\arraybackslash}X}

\usepackage{todonotes}

\definecolor{mpi-green}{RGB}{0,108,102}
\definecolor{mpi-grey}{RGB}{119,119,119}
\definecolor{mpi-lgrey}{RGB}{238,238,238}
\definecolor{mpi-lblue}{RGB}{123, 199, 208}
\definecolor{mpi-blue}{RGB}{1,147,215}
\definecolor{mpi-red}{RGB}{207,74,97}
\definecolor{mpi-lgreen}{RGB}{69,178,96}
\definecolor{mpi-orange}{RGB}{255, 186, 77}
\definecolor{mpi-purple}{RGB}{114,122,178}

\title{\LARGE \bf How to Train Your Differentiable Filter}

\author{Alina Kloss$^{1}$, Georg Martius$^{1}$ and Jeannette Bohg$^{1,2}$
\thanks{$^{1}$ Max Planck Institute for Intelligent Systems, {\tt <akloss, gmartius>@tue.mpg.de}}
\thanks{$^{2}$ Stanford University, {\tt bohg@stanford.edu}}
\thanks{The authors thank the International Max Planck Research 
School for Intelligent Systems (IMPRS-IS) for supporting Alina Kloss.}
}

\begin{document}

\maketitle
\thispagestyle{empty}
\pagestyle{empty}

\begin{abstract}
In many robotic applications, it is crucial to maintain a belief about the state
of a system, which serves as input for planning and decision making and provides
feedback during task execution. Bayesian Filtering algorithms address 
this state estimation problem, but they require models of process dynamics and 
sensory observations and the respective noise characteristics of these models.
Recently, multiple works have demonstrated that these models can be learned by 
end-to-end training through differentiable versions of recursive filtering
algorithms. In this work, we investigate the advantages of \textit{differentiable filters} 
(DFs) over both 
unstructured learning approaches and manually-tuned filtering algorithms, and 
provide practical guidance to researchers interested in applying such 
differentiable filters. For this, we implement DFs with four 
different underlying filtering algorithms and compare them in extensive 
experiments. Specifically, we (i) evaluate different implementation choices and 
training approaches, (ii) investigate how well complex models of uncertainty can
be learned in DFs, (iii) evaluate the effect of end-to-end training through DFs 
and (iv) compare the DFs among each other and to unstructured LSTM models. 
\end{abstract}

\section{Introduction}\label{intro}

In many robotic applications, it is crucial to maintain a belief 
about the state of the system over time, like tracking the location of a mobile 
robot or the pose of a manipulated object. These state estimates serve as input 
for planning and decision making and provide feedback during task execution. 
In addition to tracking the system state, it can also be desirable to estimate
the uncertainty associated with the state predictions. This information can be 
used to detect failures and enables risk-aware planning, where the robot takes
more cautious actions when its confidence in the estimated state is low
\citep{todorov-2005, ponton-2020}.

Recursive Bayesian filters are a class of algorithms that combine perception 
and prediction for probabilistic state estimation in a principled way. 
To do so, they require an observation model that relates the estimated state to 
the sensory observations and a process model that predicts how the state 
develops over time. Both have associated noise models that reflect the 
stochasticity of the underlying system and determine how much trust the filter 
places in perception and prediction. 

Formulating good observation and process models for the filters can, however, be 
difficult in many scenarios,  especially when the sensory observations are 
high-dimensional and complex, like camera images. Over the last years, deep 
learning has become the method of choice for processing such data. While 
(recurrent) neural networks can be trained to address the full state estimation 
problem directly, recent work \citep{jonschkowski-2016, haarnoja-2016, 
jonschkowski-2018, karkus-2018} showed that it is 
also possible to include data-driven models into Bayesian filters and train them 
end-to-end through the filtering algorithm. For Histogram filters 
\citep{jonschkowski-2016}, Kalman filters \citep{haarnoja-2016} and Particle 
filters \citep{jonschkowski-2018, karkus-2018}, the 
respective authors showed that such \textit{differentiable filters} (DF) 
systematically outperform unstructured neural networks like LSTMs 
\citep{hochreiter-1997}. In addition, 
the end-to-end training of the models also improved the filtering performance 
compared to using observation and process models that had been trained 
separately. 

A further interesting aspect of differentiable filters is that they allow for 
learning sophisticated models of the observation and process noise. This is 
useful because finding appropriate values for the noise models is 
often difficult and despite much research on identification methods (e.g.\ 
\cite{bavdekar-2011, valappil-2000}) they are often tuned manually 
in practice. 
To reduce the tedious tuning effort, the noise is then typically assumed to be 
uncorrelated Gaussian noise with zero mean and {\em constant} covariance. Many 
real systems are, however, better described by \textit{heteroscedastic} noise 
models, where the level of uncertainty depends on the state of the system and/or 
possible control inputs. Taking heterostochasticity of the dynamics into account 
has been demonstrated to improve filtering performance in many robotic tasks 
\citep{bauza-2017, kersting-2007}. 
\cite{haarnoja-2016} also show that learning heteroscedastic 
observation noise helps a Kalman filter dealing with occlusions during object tracking.

In this paper, we perform a thorough evaluation of differentiable filters.
Our main goals are to highlight the advantages of DFs over both unstructured 
learning approaches and manually-tuned filtering algorithms, and to provide 
guidance to practitioners interested in applying differentiable 
filtering to their problems.

To this end, we review and implement existing work on differentiable Kalman and 
Particle filters and introduce two novel variants of differentiable Unscented 
Kalman filters. 
Our implementation for TensorFlow~\citep{abadi-2015} is publicly available\footnote{\scriptsize 
\url{https://github.com/akloss/differentiable_filters}}.
In extensive experiments on three different tasks, we compare the DFs and 
evaluate different design choices for implementation and training, including 
loss functions and training sequence length. We also investigate how well the 
different filters can learn complex heteroscedastic and correlated noise models,
evaluate how end-to-end training through the DFs influences the learned models  
and compare the DFs to unstructured LSTM models. 

\section{Related Work}

\subsection{Combining Learning and Algorithms}
Integrating algorithmic structure into learning methods has been studied for 
many robotic problems, including state estimation \citep{haarnoja-2016, 
jonschkowski-2016, jonschkowski-2018, karkus-2018, ma-2020}, planning 
\citep{tamar-2016, karkus-2017, oh-2017, farquhar-2018, guez-2018} 
and control \citep{donti-2017, okada-2017, amos-2018, pereira-2018, holl-2020}.
Most notably, \cite{karkus-2019} combine multiple differentiable algorithms 
into an end-to-end trainable ``Differentiable Algorithm Network'' to address the
complete task of navigating to a goal in a previously unseen environment using
visual observations. 
Here, we focus on addressing the state estimation problem with differentiable 
implementations of Bayesian filters.

\subsection{Differentiable Bayesian Filters}

There have been few works on differentiable filters so far. \cite{haarnoja-2016}
propose the BackpropKF, a differentiable implementation of the (extended) Kalman
filter. \cite{jonschkowski-2016}
present a differentiable Histogram filter for discrete localization tasks in 
one or two dimensions and \cite{jonschkowski-2018} and \cite{karkus-2018} both 
implement
differentiable Particle filters for localization and tracking of a mobile robot. 
In the following, we focus our discussion on differentiable Kalman and Particle 
filters,
since Histogram filters as used by \cite{jonschkowski-2016} are usually not 
feasible in practice, due to the need of discretizing the complete state space.

\paragraph{Observation Model and Noise}
All three works have in common that the raw observations are processed by a 
learned neural network that can be trained end-to-end through the filter. In 
\cite{haarnoja-2016}, the network outputs a low-dimensional representation of 
the observations together with input-dependent observation noise  
(see Sec.~\ref{implementation-obs}), while in 
\cite{jonschkowski-2018, karkus-2018}, a neural network learns to predict 
the likelihood of the observations under each particle given an 
image and (in \citep{karkus-2018}) a map of the environment.

As a result, all three works use heteroscedastic observation noise, but only 
\cite{haarnoja-2016} evaluate this choice: They show that conditioning 
the observation noise on the raw image observations drastically improves filter 
performance when the tracked object can be occluded.

\paragraph{Process Model and Noise}
For predicting the next state, all three works use a given analytical process 
model. While \cite{haarnoja-2016} and \cite{karkus-2018} also assume known 
process noise, \cite{jonschkowski-2018} train a network to predict it
conditioned on the actions. The effect of learning action dependent process 
noise is, however, not evaluated.

\paragraph{Effect of End-to-End Learning}
\cite{jonschkowski-2018} compare the results of an end-to-end trained filter 
with one where the observation model and process noise were trained separately. 
The end-to-end trained variant performs better, presumably because it learns 
to overestimate the process noise. Possible differences between the learned 
observation models are not discussed. The best performance for the filter 
could be reached by first pretraining the models individually and the finetuning 
end-to-end through the filter.

\paragraph{Comparison to Unstructured Models}
All works compare their differentiable filters to LSTM models trained for 
the same task and find that including the structural priors of the filtering 
algorithm and the known process models improves performance. 
\cite{jonschkowski-2018} also evaluate a Particle filter with a learned process 
model in one experiment, which performs worse than the filter with an analytical 
process model but still beats the LSTM.
\\
\\
In contrast to the existing work on differentiable filtering, the main purpose 
of this paper is not to present a new method for solving a robotic task. 
Instead, we present a thorough evaluation of differentiable filtering and of 
implementation choices made by the aforementioned seminal works. We also 
implement two novel differentiable filters based on variants of the Unscented 
Kalman filter and 
compare the differentiable filters with different underlying Bayesian filtering
algorithms in a controlled way.

\subsection{Variational Inference}

A second line of research closely related to differentiable filters is 
variational inference in temporal state space models \citep{krishnan-2016, 
karl-2017, watter-2015, fraccaro-2017, archer-2015}. For a recent review of this
work, see \cite{girin-2020}.
In contrast to DFs, the focus of this research lies more on finding generative 
models that explain the observed data sequences and are able to generate new 
sequences. 
The representation of the underlying state of the system is often not assumed to
be known. But even though the goals are different, recent results in this field 
show that structuring the variational models similarly to Bayesian filters 
improves their performance \citep{karl-2017, fraccaro-2017, naesseth-2018, maddison-2017, anh-2018}. 

\section{Bayesian Filtering for State Estimation}\label{filtering}

Filtering refers to the problem of estimating the latent state 
$\mathbf{x}$ of a stochastic dynamic system at time step $t$ given an 
initial belief $\mathrm{bel}(\mathbf{x}_0) = p(\mathbf{x}_0)$, 
a sequence of observations 
$\mathbf{z}_{1...t}$ and actions $\mathbf{u}_{0...t-1}$. Formally, we seek 
the posterior distribution 
$\mathrm{bel}(\mathbf{x}_t) = p(\mathbf{x}_t|\mathbf{x}_{0...t-1}, \mathbf{u}_{0...t-1}, \mathbf{z}_{1...t})$. 

Bayesian Filters make the Markov assumption, i.e. that the distributions
of the future states and observations are conditionally independent from 
the history of past states and observations given the current state.
This assumption makes it possible to compute the belief at time $t$
recursively as 
\noindent
\begin{align*}
\mathrm{bel}(\mathbf{x}_t) &= 
\eta p(\mathbf{z}_t|\mathbf{x}_t) \int p(\mathbf{x}_t|\mathbf{x}_{t-1}, \mathbf{u}_{t-1}) \mathrm{bel}(\mathbf{x}_{t-1}) d \mathbf{x}_{t-1} \\
& = \eta p(\mathbf{z}_t|\mathbf{x}_t) \overline{\mathrm{bel}}(\mathbf{x}_t) 
\end{align*}
{\sloppy
where $\eta$ is a normalization factor. 
Computing $\overline{\mathrm{bel}}(\mathbf{x}_t)$ is referred to as the
\textit{prediction step} of Bayesian filters, while updating the belief with 
$p(\mathbf{z}_t|\mathbf{x}_t)$ is called \textit{(observation) update step}.

For the prediction step, the dynamics of the system is modeled by the 
\textit{process model} $f$ that describes how the state changes over time.
The observation update step uses an \textit{observation model} $h$ that generates 
observations given the current state:
\begin{align*}
\mathbf{x}_{t} &= f(\mathbf{x}_{t-1}, \mathbf{u}_{t-1}, \mathbf{q}_{t-1}) &
\mathbf{z}_{t} &= h(\mathbf{x}_{t}, \mathbf{r}_t)
\end{align*}
The random variables $\mathbf{q}$ and $\mathbf{r}$ are the process and
observation noise and capture the stochasticity of the system.

In this paper, we investigate differentiable versions of four different 
nonlinear Bayesian filtering algorithms: The Extended Kalman Filter (EKF), the 
Unscented Kalman Filter (UKF), a sampling-based variant of the UKF that 
we call Monte Carlo Unscented Kalman Filter (MCUKF) and the Particle 
Filter (PF). We briefly review these algorithms in Appendix~\ref{app:filtering}. 
}

\section{Implementation}

In this section, we describe how we embed model-learning into differentiable 
versions of the aforementioned nonlinear filtering algorithms. These 
differentiable versions will be denoted by dEKF, dUKF etc.\ in the following.

\subsection{Differentiable Filters}\label{implementation-filter}

We implement the filtering algorithms as recurrent neural network 
layers in TensorFlow. For UKF and MCUKF, this is straight-forward, since
all necessary operations are differentiable and available in TensorFlow. 

In contrast, the dEKF requires the Jacobian of the 
process model $\mathbf{F}$. TensorFlow implements a method for computing 
Jacobians, with or without vectorization. The former is fast but has a high 
memory demand, while the latter can become very slow for large batch sizes. 
Therefore, we recommend to derive the Jacobians manually where applicable.

\subsubsection{dPF}

The Particle filter is the only filter we investigate that is not fully 
differentiable: In the resampling step, a new set of particles with uniform 
weights is drawn (with replacement) from the old set according to the old 
particle weights. 
While the drawn particles can propagate gradients to their ancestors, gradient 
propagation to other old particles or to the weights of the old particle set
is disrupted \citep{jonschkowski-2018, karkus-2018, zhu-2020}. If we place the 
resampling step at the beginning of the
per-timestep computations, this only affects the gradient propagation 
through time, i.e.\ from one timestep $t+1$ to its predecessor $t$. At time $t$,
both particles and weights still receive gradient information about the 
corresponding loss at this timestep. We therefore hypothesize that the missing 
gradients through time are not problematic as long as we provide a loss at every
timestep. 

As an alternative to simply ignoring the disrupted gradients, we can also apply 
the resampling step less frequently or use soft resampling as proposed by 
\cite{karkus-2018}. We evaluate these options in 
Sec.~\ref{experiments-sim-pf-re}.
\\
\\
In addition, we investigate two alternative implementation choices for the 
dPF:
The likelihood used for updating the particle weights in the observation update 
step can be implemented either with an analytical Gaussian likelihood function 
or with a trained neural network as in \cite{jonschkowski-2018} and 
\cite{karkus-2018}. 
The learned observation likelihood is potentially more expressive than the
analytical solution and can be advantageous for problems where formulating 
the observation and sensor model is not as straight-forward as in our experiments. 
A potential drawback is 
that in contrast to the analytical solution, no explicit noise model 
or sensor network is learned. 
We compare these two options in Sec.~\ref{experiments-sim-pf-obs}.

\subsection{Observation Model}\label{implementation-obs}

In Bayesian filtering, the observation model $h(\cdot)$ is a \emph{generative} 
model that predicts observations from the state 
$\mathbf{z}_t = h(\mathbf{x}_t)$. 
In practice, it is often hard to find such models that directly 
predict the potentially high-dimensional raw sensory signals without making 
strong assumptions. 

We therefore use the method first proposed by \cite{haarnoja-2016} and train a 
\emph{discriminative} neural network $n_s$ with parameters $\mathbf{w}_s$ to 
preprocess the raw sensory data $\mathbf{D}$ and create a more compact 
representation of the observations $\mathbf{z} = n_s(\mathbf{D}, \mathbf{w}_s)$. 
This network can be seen as a virtual sensor, and we thus call it 
\textit{sensor network}. In addition to $\mathbf{z}_t$, the sensor network can 
also predict the heteroscedastic observation noise covariance matrix 
$\mathbf{R}_t$ (see Sec.~\ref{implementation-noise}) for the current input 
$\mathbf{D}_t$. 

In our experiments, $\mathbf{z}$ contains a subset of the state vector 
$\mathbf{x}$. The actual observation model $h(\mathbf{x})$ thus reduces 
to a simple linear selection matrix of the observable components, which we 
provide to the DFs.

\subsection{Process Model}\label{implementation-proc}

Depending on the user's knowledge about the system, the process model $f(\cdot)$
for the prediction step can be implemented using a known analytical model or a 
neural network $n_p(\cdot)$ with weights $\mathbf{w}_p$. When using neural 
networks, we train $n_p(\cdot)$ to output the change from the last state 
$n_p(\mathbf{x}_{t}, \mathbf{u}_t, \mathbf{w}_p) = \Delta \mathbf{x}_t$ such 
that $\mathbf{x}_{t+1} = \mathbf{x}_{t} + \Delta \mathbf{x}_{t}$. This form 
ensures stable gradients between timesteps 
(since $\frac{\partial \mathbf{x}_{t+1}}{\partial \mathbf{x}_{t}} = 1 + 
\frac{\partial p}{\partial \mathbf{x}_{t}}$) and provides a reasonable 
initialization of the process model close to identity.

\subsection{Noise Models}\label{implementation-noise}

For learning the observation and process noise, we consider two different 
conditions: constant and heteroscedastic. In both cases, we assume that the  
process and observation noise at time $t$ can be described by zero-mean Gaussian
distributions with covariance matrices $\mathbf{Q}_t$ and $\mathbf{R}_t$. 

A common assumption in state-space modeling is that $\mathbf{Q}_t$ and 
$\mathbf{R}_t$ are diagonal matrices, but we can also use full covariance 
matrices to model correlated noise. In this case, we follow \cite{haarnoja-2016}
and train the noise models to output upper-triangular matrices $\mathbf{L}_t$, 
such that e.g.\  $\mathbf{Q}_t = \mathbf{L}_t\mathbf{L}_t^T$. This form ensures 
that the resulting matrices are positive definite.

For constant noise, the filters directly learn the diagonal or triangular 
elements of $\mathbf{Q}$ and $\mathbf{R}$. In the heteroscedastic case, 
$\mathbf{Q}_t$ is predicted from the current state $\mathbf{x}_t$ and (if 
available) the control input $\mathbf{u}_t$ by a neural network 
$n_q(\mathbf{x}_{t}, \mathbf{u}_{t}, \mathbf{w}_q)$ with weights $\mathbf{w}_q$.
In dUKF, dMCUKF and dPF, $n_q(\cdot)$ outputs separate $\mathbf{Q}^i$ for each 
sigma point/particle and $\mathbf{Q}_t$ is computed as their weighted mean. The 
heteroscedastic observation noise covariance matrix $\mathbf{R}_t$ is an 
additional output of the sensor model $n_s(\mathbf{D}_t, \mathbf{w}_s)$.

We initialize the diagonals $\mathbf{Q}_t$ and $\mathbf{R}_t$ close to 
given target values by adding a trainable bias variable to the output of the 
noise models. To prevent numerical instabilities, we also add a small fixed 
bias to the diagonals as a lower bound for the predicted 
noise. 

\subsection{Loss Function}\label{implementation-train}

For training the filters, we always assume that we have access to the ground 
truth trajectory of the state $\mathbf{x}^l_{t=0...T}$. 
In our experiments, we test the two different loss functions used in related 
work: The first, used by \cite{karkus-2018} is simply the mean squared error 
(MSE) between the mean of the belief and true state at each timestep:
\begin{equation}
L_{\mathrm{MSE}} = \frac{1}{T}\sum_{t=0}^T (\mathbf{x}^l_t - 
\boldsymbol{\mu}_t)^T(\mathbf{x}^l_t - \boldsymbol{\mu}_t).\label{eq:mse}
\end{equation}
For the dPF, we compute $\mathbf{\mu}$ as the weighted mean of the particles.

The second loss function, used by \cite{haarnoja-2016} and \cite{jonschkowski-2018}, is 
the negative log likelihood (NLL) of the true state under the predicted 
distribution of the belief. 
In dEKF, dUKF and dMCUKF, the belief is represented by a Gaussian distribution 
with mean $\boldsymbol{\mu}_t$ and covariance $\boldsymbol{\Sigma}_t$ and the 
negative log likelihood is computed as
\begin{equation} 
L_{\mathrm{NLL}} = \frac{1}{2T}\sum_{t=0}^T \log(|\boldsymbol{\Sigma}_t|) + 
(\mathbf{x}^l_t - \boldsymbol{\mu}_t)^T\boldsymbol{\Sigma}_t^{-1}(\mathbf{x}^l_t - \boldsymbol{\mu}_t).\label{eq:like}
\end{equation}

The dPF represents its belief using the particles 
$\boldsymbol{\chi}_i \in \boldsymbol{\Chi}$ and their weights $\pi_i$. We 
consider two alternative ways of calculating the NLL for training the dPF: The 
first is to represent the belief by fitting a single Gaussian to the particles, 
with $\boldsymbol{\mu} = \sum_{i=0}^N \pi_i \boldsymbol{\chi}_i$ and
$\boldsymbol{\Sigma} = \sum_{i=0}^N \pi_i (\boldsymbol{\chi}_i - \boldsymbol{\mu})(\boldsymbol{\chi}_i - \boldsymbol{\mu})^T$
and then apply Eq.~\ref{eq:like}. We refer to this variant as dPF-G.

However, this is only a good representation of the belief if the distribution 
of the particles is unimodal. To better reflect the potential multimodality of 
the particle distribution, the belief can also be represented with a Gaussian 
Mixture Model (GMM) as proposed by \cite{jonschkowski-2018}. Every particle 
contributes a separate Gaussian $N_i(\boldsymbol{\chi}^i, \boldsymbol{\Sigma})$ 
in the GMM and the mixture weights are the particle weights. The drawback of this 
approach is that the fixed covariance $\boldsymbol{\Sigma}$ of the individual 
distributions is an additional tuning parameter for the filter. We call this 
version dPF-M and calculate the negative log likelihood with
\begin{equation} 
L_{\mathrm{NLL}} = \frac{1}{T}\sum_{t=0}^T \log \sum_{i=0}^{|\boldsymbol{\Chi}|} 
\frac{\pi^i}{\sqrt{|\boldsymbol{\Sigma}|}}
\exp(\mathbf{x}^l_t - \boldsymbol{\chi}^{i}_t)^T\boldsymbol{\Sigma}^{-1}(\mathbf{x}^l_t - \boldsymbol{\chi}^i_t)\label{eq:like-gmm}
\end{equation} 

\section{Experimental Setup}\label{experiments}

In the following, we will evaluate the DFs on three different filtering 
problems. We start with a simple simulation setting that gives us full control 
over parameters of the system such as the true process noise 
(Sec.~\ref{experiments-sim}). 
In Sections~\ref{experiments-kitti} and \ref{experiments-pushing}, we then 
study the performance of the DFs on two real-robot tasks: The first is the KITTI
Visual Odometry problem, where the filters are used to track the position and
heading of a moving car given only RGB images. The second is planar pushing, 
where the filters track the pose of an object while a robot performs a series of
pushes.

Unless stated otherwise, we will train the DFs end-to-end for 15 epochs using 
the Adam optimizer \citep{kingma-2015} and select the model state at the training step 
with the best validation loss for evaluation. We also evaluate different learning 
rates for all DFs.
During training, the initial state is 
perturbed with noise sampled from a Normal distribution 
$N_{\mathrm{init}}(0, \boldsymbol{\Sigma}_{\mathrm{init}})$. 
For testing, we evaluate all DFs with the true initial state as well as 
with few fixed perturbations (sampled from $N_{\mathrm{init}}$) and average the results. 

More detailed information about the experimental conditions as well as extended 
results can be found in Appendix~\ref{app:sim}-\ref{app:push}.

\section{Simulated Disc Tracking}\label{experiments-sim}

\begin{figure}
\centering
\includegraphics[width=0.45\linewidth]{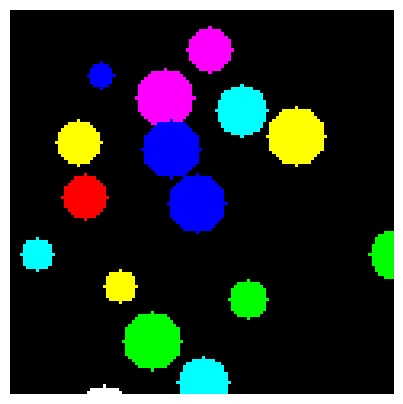}
\includegraphics[width=0.45\linewidth]{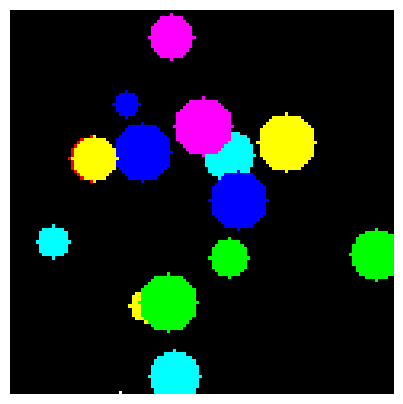}
\caption{Two sequential observations from our simulated tracking task. The 
filters need to track the red disc, which can be occluded by the 
other discs or leave the image temporarily.}\label{fig-observations}
\end{figure}

We first evaluate the DFs in a simulated environment similar to the one in 
\cite{haarnoja-2016}: the task is to track a red disc moving among varying 
numbers of distractor discs, as shown in Figure~\ref{fig-observations}. The state 
consists of the position $\mathbf{p}$ and linear velocity $\mathbf{v}$ of the 
red disc.

The dynamics model that we use for generating the training data is 
\begin{align*}
\mathbf{p}_{t+1} &= \mathbf{p}_t + \mathbf{v}_t + \mathbf{q}_{p,t} \\
\mathbf{v}_{t+1} &= \mathbf{v}_t - f_p \mathbf{p}_t 
- f_d \mathbf{v}_{t}^2 \mathrm{sgn}(\mathbf{v}_t) + \mathbf{q}_{v, t}
\end{align*}
The velocity update contains a force that pulls the discs towards the origin 
($f_p = 0.05$) and a drag force that prevents too high velocities 
($f_d = 0.0075$). $\mathbf{q}$ represents the Gaussian process noise and
$\mathrm{sgn}(x)$ returns the sign of $x$ or 0 if $x=0$.

The sensor network receives the current image at each step, from which it can 
estimate the position but not the velocity of the target. As we do not model 
collisions, the red disc can be occluded by the distractors or leave the image 
temporarily. 

\subsection{Data}\label{sim-data}

We create multiple datasets with varying numbers of distractors, different 
levels of constant process noise for the disc position and constant or 
heteroscedastic process noise for the disc velocity. All datasets contain 
2400 sequences for training, 300 validation sequences and 303 sequences for 
testing. The sequences have 50 steps and the colors and sizes of the distractors 
are drawn randomly for each sequence. 

\subsection{Filter Implementation and Parameters}\label{sim-imp}

We first evaluated different design choices and filter-specific parameters for 
the DFs to find settings that perform well and increase the stability of the 
filters during training. 
For detailed information about the experiments and results,
please refer to Appendix~\ref{app:sim-imp}.

\subsubsection{dUKF}\label{sim-ukf}

The dUKF has three filter-specific scaling parameters, $\alpha$, $\kappa$ and 
$\beta$. $\alpha$ and $\kappa$ determine 
how far from the mean of the belief the sigma points are placed and how the mean
is weighted in comparison to the other sigma points. $\beta$ only affects the
weight of the central sigma point when computing the covariance of the 
transformed distribution. 

We evaluated different parameter settings but found no significant differences 
between them. In all following experiments, we use $\alpha=1$, $\kappa=0.5$
and $\beta=0$. In general, we recommend values for which  
$\lambda = \alpha^2(\kappa + n) - n$ is a small positive number, 
so that the sigma points are not spread out too far and the central sigma 
point is not weighted negatively (which happens for negative $\lambda$). 
See Appendix~\ref{app:filtering:UKF} for a more detailed
explanation.  

\subsubsection{dMCUKF}\label{sim-mcukf}

In contrast to the dUKF, the dMCUKF simply samples pseudo sigma points from
the current belief. Its only parameter thus is the number $N$ of sampled points 
during training and testing. 

We trained the dMCUKF with $N \in \{5, 10, 50, 100, 500\}$ and evaluated 
with 500 pseudo sigma points.
The results show that as few as ten sigma points are enough for training 
the dMCUKF relatively successfully. The best results are obtained with 100 sigma
points and using more does not reliably increase the performance. 

In the following, we use 100 points for training and 500 
for testing. More complex problems with higher-dimensional states could, 
however, require more sigma points.

\subsubsection{dPF: Belief Representation}\label{experiments-sim-pf-bel}

When training the dPF on $L_{\mathrm{NLL}}$, we have to choose how to represent the 
belief of the filter for computing the likelihood (see 
Sec.~\ref{implementation-train}). 
We investigate using a single Gaussian (dPF-G) or a Gaussian Mixture Model 
(dPF-M). For the dPF-M, the covariance $\boldsymbol{\Sigma}$ of the single 
Gaussians in the Mixture Model is an additional parameter that has to be tuned. 

As our test scenario does not require tracking multiple hypotheses, the 
representation by a single Gaussian in dPF-G should be accurate for this task.
Nonetheless, we find that the dPF-G  performs much worse than the dPF-M. 
This could either mean that Eq.~\ref{eq:like-gmm} facilitates training or that 
approximating the belief with a single Gaussian removes useful information even
when the task does not obviously require tracking multiple hypotheses. 
Interestingly, when using a learned observation update, this effect is not 
noticeable, which suggests that the first hypothesis is correct.
In the following, we only report results for the dPF-M. Results for dPF-G
can be found in the Appendix.

For the dPF-M, $\boldsymbol{\Sigma} = 0.25 \mathbf{I}_4$ ($\mathbf{I}_4$ denotes 
an identity matrix with 4 rows and columns) resulted in the best tracking errors, 
but the best NLL was achieved with $\boldsymbol{\Sigma} = \mathbf{I}_4$. We thus use 
$\boldsymbol{\Sigma} = \mathbf{I}_4$ for the dPF-M in all following experiments. 
It is, however, possible that different tasks could require different 
settings.

\subsubsection{dPF: Observation Update}\label{experiments-sim-pf-obs}

As mentioned before, the likelihood for the observation update step of the dPF
can be implemented with an analytical Gaussian likelihood function (dPF-(G/M))
or with a neural network (dPF-(G/M)-lrn).

Our experiments showed that using a learned likelihood function for updating the
particle weights can improve both tracking error and NLL of the dPF 
significantly. We attribute this mainly to the fact that the learned update 
relaxes some of the assumptions encoded in the particle filter:
With the analytical version, we restrict the filter to use
additive Gaussian noise that is either constant or depends only on the raw 
sensory observations. The learned update, in contrast, enforces no functional 
form of the noise model. In addition, the noise can depend not only on the raw 
sensory data, but also on the observable components of the particle states. This
means that the learned observation update is potentially much more expressive 
than the analytical one, which pays off when the Gaussian assumption made
by the other filtering algorithms does not hold.

While learning the observation update improves the performance of the dPF,
we will still use the analytical variant in most of the following evaluations. 
The main reason for this is that the analytical observation update has explicit
models for the sensor network and observation noise.
This facilitates comparing between the dPF and the other DF variants and gives
us control over the form of the learned observation noise.

\subsubsection{dPF: Resampling}\label{experiments-sim-pf-re}

The resampling step of the particle filter discards particles with low weights 
and prevents particle depletion. It may, however, be disadvantageous during 
training since it is not fully differentiable. \cite{karkus-2018} proposed 
soft resampling, where the resampling distribution is traded off with a uniform 
distribution to enable gradient flow between the weights of the old and new 
particles. This trade-off is controlled by a parameter $\alpha_{\mathrm{re}} \in [0, 1]$.
The higher $\alpha_{\mathrm{re}}$, the more weight is put on the uniform distribution. An
alternative to soft resampling is to not resample at every timestep.

We tested the dPF-M with different values of $\alpha_{\mathrm{re}}$ and  
when resampling every 1, 2, 5 or 10 steps and found that resampling frequently 
generally improves the filter performance. Soft resampling also did not have 
much of a positive effect in our experiments, presumably because higher values 
of $\alpha_{\mathrm{re}}$ decrease the effectiveness of the resampling step. In the 
following, we use $\alpha_{\mathrm{re}}=0.05$ and resample at every timestep.

\subsubsection{dPF: Number of Particles}\label{experiments-sim-pf-num}

Finally, the user also has to decide how many particles to use during training 
and testing. As for the dMCUKF, we trained the dPF-M with 
$N \in \{5, 10, 50, 100, 500\}$. 
The results were very similar to dMCUKF and we also use 100 particles 
during training and 500 particles for testing.

\begin{figure*}[ht!]
\centering
\begin{tikzpicture}
	\begin{groupplot}[
		group style={
            group size=3 by 1,
            horizontal sep=0.75cm,
            vertical sep=0.35cm},
        enlarge x limits=0.2,
        width=7cm,
        height=4.cm,
        xtick={data},
        ylabel style={yshift=0.05cm},
        ybar=0.pt,
        symbolic x coords={dEKF, dUKF, dMCUKF, dPF-M},
        xticklabel style={rotate=15},
        ]
	\nextgroupplot[title={tracking RMSE}, 
    			   bar width=0.275cm, 
    			   ymin=0, ymax=15]
	  \addplot[mpi-red!70!black, fill=mpi-red!65, 
	  		   error bars/.cd, y dir=both, y explicit, error bar style={line width=0.9pt}
	  		   ] table
        [x=filter, y=error, y error=error-std] {img/sim/loss_dist}; 
      \addplot[mpi-blue!70!black, fill=mpi-blue!65, 
	  		   error bars/.cd, y dir=both, y explicit, error bar style={line width=0.9pt}
	  		   ] table
        [x=filter, y=mix, y error=mix-std] {img/sim/loss_dist}; 
      \addplot[mpi-orange!70!black, fill=mpi-orange!65, 
	  		   error bars/.cd, y dir=both, y explicit, error bar style={line width=0.9pt}
	  		   ] table
        [x=filter, y=like, y error=like-std] {img/sim/loss_dist}; 

	\nextgroupplot[title={observation RMSE}, width=6cm, enlarge x limits=0.2,
    			   bar width=0.275cm, ymin=0, ymax=15,
    			   legend style={at={(0.5,-0.35)}},
			       legend columns=-1, legend entries={$L_{\mathrm{MSE}}$, $L_{\mathrm{mix}}$, $L_{\mathrm{NLL}}$}]
	  \addplot[mpi-red!70!black, fill=mpi-red!65, 
	  		   error bars/.cd, y dir=both, y explicit, error bar style={line width=0.9pt}
	  		   ] table
        [x=filter, y=error, y error=error-std] {img/sim/loss_obs}; 
      \addplot[mpi-blue!70!black, fill=mpi-blue!65, 
	  		   error bars/.cd, y dir=both, y explicit, error bar style={line width=0.9pt}
	  		   ] table
        [x=filter, y=mix, y error=mix-std] {img/sim/loss_obs}; 
      \addplot[mpi-orange!70!black, fill=mpi-orange!65, 
	  		   error bars/.cd, y dir=both, y explicit, error bar style={line width=0.9pt}
	  		   ] table
        [x=filter, y=like, y error=like-std] {img/sim/loss_obs};

    \nextgroupplot[ 
    			    title={-log likelihood},
    			   bar width=0.275cm, ymin=0, ymax=30]
	  \addplot[mpi-red!70!black, fill=mpi-red!65, 
	  		   error bars/.cd, y dir=both, y explicit, error bar style={line width=0.9pt}
	  		   ] table
        [x=filter, y=error, y error=error-std] {img/sim/loss_like}; 
      \addplot[mpi-blue!70!black, fill=mpi-blue!65, 
	  		   error bars/.cd, y dir=both, y explicit, error bar style={line width=0.9pt}
	  		   ] table
        [x=filter, y=mix, y error=mix-std] {img/sim/loss_like}; 
      \addplot[mpi-orange!70!black, fill=mpi-orange!65, 
	  		   error bars/.cd, y dir=both, y explicit, error bar style={line width=0.9pt}
	  		   ] table
        [x=filter, y=like, y error=like-std] {img/sim/loss_like}; 
        
    \end{groupplot}
\end{tikzpicture}
\caption{Results on disc tracking: Performance of the DFs when trained with 
different loss functions 
$L_{\mathrm{MSE}}$, $L_{\mathrm{NLL}}$ or $L_{\mathrm{mix}}$ averaged over all 
steps in the trajectory. The first plot shows the RSME of the estimated state while
the second shows the difference between the output of the sensor network and
the corresponding ground truth state components. The negative log likelihood is 
computed as shown in Eq.~\ref{eq:like} or Eq.~\ref{eq:like-gmm} for the dPF-M.}\label{fig-sim-loss}
\end{figure*}
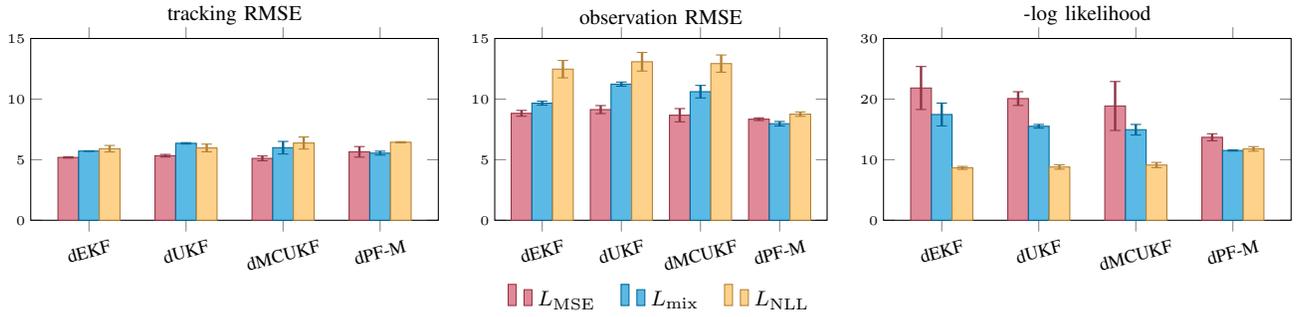

\begin{figure*}[t!]
\centering
\begin{tikzpicture}
	\begin{groupplot}[
		group style={
            group size=2 by 1, 
            horizontal sep=1.25cm},
        enlarge x limits=0.1,
        ylabel style={yshift=0.05cm},
        xlabel style={yshift=-0.1cm},
        width=9.2cm,
        height=4.cm,
        xtick={data},
        ybar=0.5pt,
        symbolic x coords={1, 2, 5, 10, 25, 50}]
        
	\nextgroupplot[ylabel style={align=center}, ylabel={RMSE}, 
				   xlabel={training sequence length},  
				   bar width=0.25cm, ymax=25, ymin=0]
	  \addplot[mpi-red!70!black, fill=mpi-red!65, 
	  		   error bars/.cd, y dir=both, y explicit, error bar style={line width=0.9pt}
	  		   ] table
        [x=sl, y=ekf-dist, y error=ekf-dist-std] {img/sim/sl};  
      \addplot[mpi-blue!70!black, fill=mpi-blue!65, 
	  		    error bars/.cd, y dir=both, y explicit, error bar style={line width=0.9pt}
	  		   ] table
        [x=sl, y=ukf-dist, y error=ukf-dist-std] {img/sim/sl}; 
      \addplot[mpi-orange!70!black, fill=mpi-orange!65, 
	  		   error bars/.cd, y dir=both, y explicit, error bar style={line width=0.9pt}
	  		   ] table
        [x=sl, y=mcukf-dist, y error=mcukf-dist-std] {img/sim/sl}; 
      \addplot[mpi-lgreen!70!black, fill=mpi-lgreen!65, 
	  		   error bars/.cd, y dir=both, y explicit, error bar style={line width=0.9pt}
	  		   ] table
        [x=sl, y=pf-dist, y error=pf-dist-std] {img/sim/sl};

    \nextgroupplot[ylabel style={align=center}, ylabel={-log likelihood}, xlabel={training sequence length},  
    			   bar width=0.25cm, 
    			   legend style={at={(-0.1,-0.35)}},
			       legend columns=-1, legend entries={dEKF, dUKF, dMCUKF, dPF-M},
			       ymax=25, ymin=0]
	  \addplot[mpi-red!70!black, fill=mpi-red!65, 
	  		   error bars/.cd, y dir=both, y explicit, error bar style={line width=0.9pt}
	  		   ] table
        [x=sl, y=ekf-like, y error=ekf-like-std] {img/sim/sl};  
      \addplot[mpi-blue!70!black, fill=mpi-blue!65, 
	  		   error bars/.cd, y dir=both, y explicit, error bar style={line width=0.9pt}
	  		   ] table
        [x=sl, y=ukf-like, y error=ukf-like-std] {img/sim/sl}; 
      \addplot[mpi-orange!70!black, fill=mpi-orange!65, 
	  		   error bars/.cd, y dir=both, y explicit, error bar style={line width=0.9pt}
	  		   ] table
        [x=sl, y=mcukf-like, y error=mcukf-like-std] {img/sim/sl};
      \addplot[mpi-lgreen!70!black, fill=mpi-lgreen!65, 
	  		    error bars/.cd, y dir=both, y explicit, error bar style={line width=0.9pt}
	  		   ] table
        [x=sl, y=pf-like, y error=pf-like-std] {img/sim/sl}; 

    \end{groupplot}
\end{tikzpicture}
\caption{Results on disc tracking: Performance of the DFs trained with different 
sequence lengths. The cut-off NLL values for sequence length 1 are 47.4$\pm$3.9
for the dUKF and 79.0$\pm$	2.7 for the dPF-M.}\label{fig:sl}
\end{figure*}
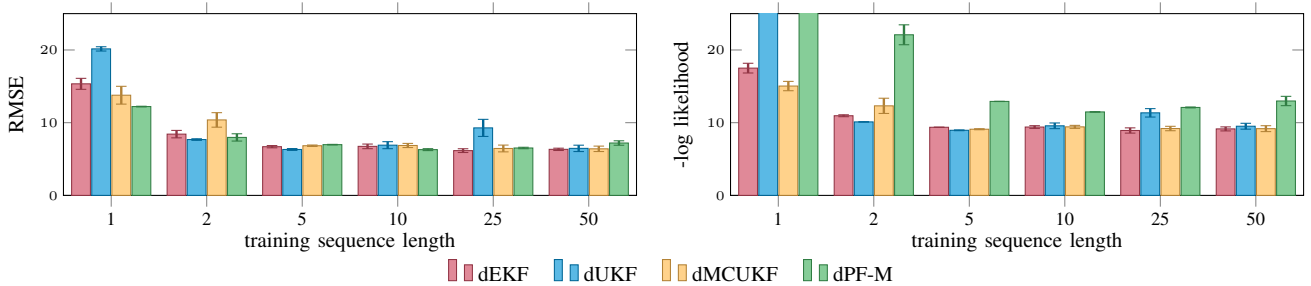

\subsection{Loss Function}\label{experiments-sim-loss}

In this experiment we compare the different loss functions introduced in 
Sec.~\ref{implementation-train}, as well as a combination of the two 
$L_{\mathrm{\mathrm{mix}}} = 0.5(L_{\mathrm{MSE}} + L_{\mathrm{NLL}})$. Our 
hypothesis is that $L_{\mathrm{NLL}}$ is better 
suited for learning noise models, since it requires predicting the uncertainty 
about the state, while $L_{\mathrm{MSE}}$ only optimizes the tracking performance.

\paragraph{Experiment}

We use a dataset with 15 distractors and constant process noise 
($\sigma_{q_p} = 0.1$, $\sigma_{q_v} = 2$). 
The filters learn the sensor and process model 
as well as heteroscedastic observation noise and constant process noise models. 

\paragraph{Results}
As expected, training on $L_{\mathrm{NLL}}$ leads to much better likelihoods scores than 
training on $L_{\mathrm{MSE}}$ for all DFs, see Fig.~\ref{fig-sim-loss}. The best 
tracking errors on the other hand are reached with $L_{\mathrm{MSE}}$, as well as more 
precise sensor models. 

For judging the quality of a DF, both NLL and tracking error should be 
taken into account: While a low RMSE is important for all tasks that use the 
state estimate, a good likelihood means that the uncertainty about the state is 
communicated correctly, which enables e.g.\ risk-aware planning and failure
detection.

The combined loss $L_{\mathrm{mix}}$ trades off these two objectives during training. It 
does, however, not outperform the single losses in their respective objective. 
A possible explanation is that they can result in opposing gradients:
All DFs tend to overestimate the process noise when trained only on $L_{\mathrm{MSE}}$. 
This lowers the tracking error by giving more weight to the observations in 
dEKF, dUKF and dMCUKF and allowing more exploration in the dPF. But it also 
results in a higher uncertainty about the state, which is undesirable when 
optimizing the likelihood. 

We generally recommend using $L_{\mathrm{NLL}}$  during training to ensure learning 
accurate noise models. If learning the process and sensor model does not work 
well, $L_{\mathrm{NLL}}$ can either be combined with $L_{\mathrm{MSE}}$ or the models can be 
pretrained.

\subsection{Training Sequence Length}\label{experiments-sim-sl}

\cite{karkus-2018} evaluated training their dPF on sequences of length 
$k \in \{1, 2, 4\}$ and found that using more steps improved results. Here, we 
want to test if increasing the sequence length even further is beneficial. 
However, longer training sequences also mean longer training times (or more 
memory consumption). We thus aim to find a value for $k$ with a good trade off 
between training speed and model performance.
 
\paragraph{Experiment}

We evaluate the DFs on a dataset with 15 distractors and constant process noise 
($\sigma_{q_p} = 0.1$, $\sigma_{q_v} = 2$).
The filters learn the sensor and process model as well as heteroscedastic 
observation noise and constant process noise models. We train using $L_{\mathrm{NLL}}$ 
on sequence lengths $k \in \{1, 2, 5, 10, 25, 50\}$ while keeping the total 
number of examples per batch (steps $\times$ batch size) constant.

\paragraph{Results}

Our results in Figure~\ref{fig:sl} show that all filters benefit from longer 
training sequences much more than the results in \cite{karkus-2018} indicated. 
However, while only one time step is clearly too little, returns diminish after 
around ten steps. 

Why are longer training sequences helpful? One issue with short sequences is 
that we use noisy initial states during training. This reflects real-world 
conditions, but the noisy inputs hinder learning the process model. On longer 
sequences, the observation updates can improve the state estimate and thus 
provide more accurate input values.

We repeated the experiment without perturbing the initial state, but the results
with $k \in \{1,2\}$ got even worse: Since the DFs could now learn accurate 
process models, they did not need the observations to achieve a low training 
loss and thus did not learn a proper sensor model. On the longer test sequences,
however, even small errors from the noisy dynamics accumulate over time if they 
are not corrected by the observations.

To summarize, longer sequences are beneficial for training DFs, because they 
demonstrate error accumulation during filtering and allow for convergence of the
state estimate when the initial state is noisy. However, performance eventually
saturates and increasing $k$ also increased our training times. We therefore 
chose $k=10$ for all experiments, which provides a good trade-off between 
training speed and performance.

\subsection{Learning Noise Models}\label{sim-noise}

The following experiments analyze how well complex models of the process 
and observation noise can be learned through the filters and how much this 
improves the filter performance. 
To isolate the effect of the noise models, we use a fixed, pretrained sensor 
model and the true analytical process model, such that only the noise models are
trained. We initialize $\mathbf{Q}$ and $\mathbf{R}$ with 
$\mathbf{Q} = \mathbf{I}_4$ and $\mathbf{R} = 100 \mathbf{I}_2$. All DFs are 
trained on $L_{\mathrm{NLL}}$. 

Appendix~\ref{app:sim-noise} contains extended experimental results on additional 
datasetsas well as data for the dPF-G.

\subsubsection{Heteroscedastic Observation Noise}\label{sim-noise-obs}

We first test if learning more complex, heteroscedastic observation noise 
models improves the performance of the filters as compared to learning constant 
noise models. For this, we compare DFs that learn constant or heteroscedastic 
observation noise (the process noise is constant) on a 
dataset with constant process noise ($\sigma_{q_p} = 3$, $\sigma_{q_v} = 2$)
and 30 distractors. 

To measure how well the predicted observation noise reflects the visibility of 
the target disc, we compute the correlation coefficient between the predicted 
$\mathbf{R}$ and the number of visible target pixels. 
We also evaluate the similarity between the learned and the true process noise 
model using the Bhattacharyya distance. 

\begin{table}
\caption{Results for disc tracking: End-to-end learning of the noise models 
through the DFs on datasets
with 30 distractors and different levels of process noise. While 
$\mathbf{Q}$ is always constant, we evaluate learning constant (const.) or 
heteroscedastic (hetero) observation noise $\mathbf{R}$. We show the tracking 
error (RMSE), negative log likelihood (NLL), the correlation coefficient between 
predicted $\mathbf{R}$ and the number of visible pixels of the target disc 
(corr.) and the Bhattacharyya distance between true and learned process 
noise model ($D_{\mathbf{Q}}$). The best results per DF are highlighted in bold.
}\label{tab:obs_noise}
\footnotesize
\centering
\begin{tabularx}{\linewidth}{X X  Y Y Y Y} 
\toprule    
  & R &  RMSE  & NLL   & corr. & $D_{\mathbf{Q}}$  \\
\midrule
\multirow{2}*{{dEKF}} 
 & const.  &   16.2 & 14.0 & - &  0.121  \\
 & hetero. &   \textbf{8.8} & \textbf{10.7} & -0.78 & \textbf{0.002}  \\
\cmidrule{2-6}
\multirow{2}*{{dUKF}} 
  & const.  &   16.8 & 14.1 & - &  0.161 \\
  & hetero. &   \textbf{8.8} & \textbf{10.7} & -0.78 & \textbf{0.013}  \\
\cmidrule{2-6}
\multirow{2}*{{dMCUKF}} 
  & const.  &   16.7 & 14.1 & - & 0.152   \\
  & hetero. &   \textbf{9.0} & \textbf{10.9} & -0.78 & \textbf{0.006} \\
\cmidrule{2-6}
\multirow{2}*{{dPF-M}} 
& const.  &   16.1 & 34.3 & - & 0.435 \\
& hetero. &   \textbf{9.6} & \textbf{20.8} & -0.77 & \textbf{0.280} \\
\bottomrule
\end{tabularx}
\end{table}

\paragraph{Results}

Results are shown in Table~\ref{tab:obs_noise}. 
When learning constant observation noise, all DFs perform relatively bad in 
terms of the tracking error. Upon inspection, we find that all filters
learn a very high $\mathbf{R}$ and thus mostly rely on the process model for 
their prediction. For example, the dEKF predicts $\sigma_{r_p} = 25.4$.
This is expected, since trusting the observations would result in wrong updates 
to the mean state estimate when the target disc is occluded.

Like \cite{haarnoja-2016}, we find that heteroscedastic observation noise 
significantly improves the tracking performance of all DFs (except for the 
dPF-M). The strong negative correlation between $\mathbf{R}$ and the visible 
disc pixels shows that the DFs correctly predict higher uncertainty when the 
target is occluded. For example, the dEKF predicts values as low as
$\sigma_{r_p} = 0.9$ when the disc is perfectly visible and as high as
$\sigma_{r_p} = 29.3$ when it is fully occluded.

Finally, all DFs learn values of $\mathbf{Q}$ that are close to the ground 
truth. For dEKF, dUKF and dMCUKF, 
the results improve significantly when 
heteroscedastic observation noise is learned. This could be because the worse 
tracking performance with constant observation noise impedes learning an 
accurate process model and thus requires higher process noise.

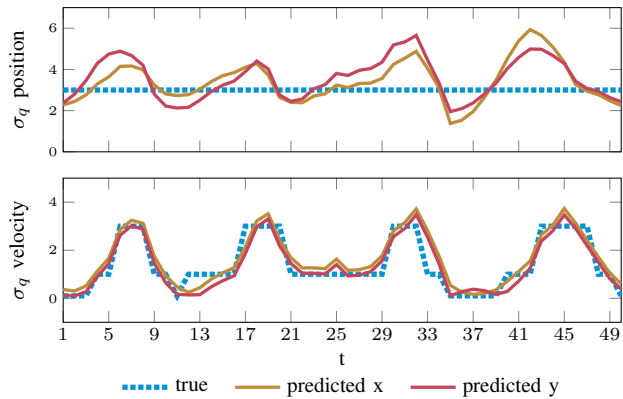
\begin{figure}
\centering
\begin{tikzpicture}
	\begin{groupplot}[
		group style={
            group size=1 by 2, 
            vertical sep=0.35cm},
        width=9cm,
        height=3.5cm,
        xtick={data},
        ymax=40, ymin=25,
        xtick={1,5,...,50},
        ]
        
	\nextgroupplot[ylabel style={align=center}, ylabel={$\sigma_{q}$ position}, 
	               ymax=7, ymin=0, xticklabel=\empty]
	  \addplot[mpi-blue, densely dotted, line width=2pt] table [y=qx] {img/sim/qh_ekf}; 
      \addplot[mpi-orange!75!black, line width=1.25pt] table [y=qx_p] {img/sim/qh_ekf}; 
      \addplot[mpi-red!95!black, line width=1.25pt] table [y=qy_p] {img/sim/qh_ekf}; 
      
    \nextgroupplot[ylabel style={align=center}, ylabel={$\sigma_{q}$ velocity}, 
    			   xlabel={t}, ymax=5, ymin=-1,
 				   legend style={at={(0.5,-0.35)}},
			        legend columns=-1, 
			        legend entries={true, predicted x, predicted y}]]
	  \addplot[mpi-blue, densely dotted, line width=2pt] table [y=qvx] {img/sim/qh_ekf}; 
      \addplot[mpi-orange!75!black, line width=1.25pt] table [y=qvx_p] {img/sim/qh_ekf}; 
      \addplot[mpi-red!95!black, line width=1.25pt] table [y=qvy_p] {img/sim/qh_ekf};
    \end{groupplot}
\end{tikzpicture}
\caption{Predicted and true process noise from the dEKF over one test sequence of
the disc tracking task. Our 
model predicts separate values for the x and y-coordinates of position and 
velocity, but the ground truth process noise has the same $\sigma$ for both 
coordinates.}\label{fig:hetero-q}
\end{figure}

\subsubsection{Heteroscedastic Process Noise}\label{sim-noise-proc}

\begin{table*}[t!]
\footnotesize
\centering
\caption{Results on disc tracking: End-to-end learning of constant or 
heteroscedastic process noise
$\mathbf{Q}$ on datasets with 30 distractors and heteroscedastic 
or constant ($\sigma_{q_p}=3.0$, $\sigma_{q_v}=2.0$) process noise. 
$D_{\mathbf{Q}}$ is the Bhattacharyya distance between true and learned process 
noise. The best results per DF are highlighted in bold.
}\label{tab:noise_qh}
\begin{tabular}{l l l c c c  l c c c } 
\toprule
  &   & & \multicolumn{3}{c}{heteroscedastic noise} & & \multicolumn{3}{c}{constant noise}  \\
  & Q & &  RMSE & NLL & $D_{\mathbf{Q}}$            & & RMSE & NLL & $D_{\mathbf{Q}}$   \\
\cmidrule{2-10}
\multirow{2}*{{dEKF}} 
  & const. & &  8.09 & 11.620 & 0.879   & &  8.80 & 10.687 & \textbf{0.002}   \\
  &	hetero.	& &  \textbf{7.36} & \textbf{11.289} & \textbf{0.402}  & &  \textbf{8.77} & \textbf{10.684} & 0.033   \\
\cmidrule{2-10}
\multirow{2}*{{dUKF}}
  & const.  & & 7.85 & 11.318 & 0.874  &  &  8.80 & 10.743 & \textbf{0.013}   \\
  & hetero. & & \textbf{7.60} & \textbf{11.167} & \textbf{0.391}   &&  \textbf{8.68} & \textbf{10.727} & 0.030  \\
\cmidrule{2-10}
\multirow{2}*{{dMCUKF}}
  & const.  & & 8.13 & 11.493 & 0.891  &  & 8.98 & 10.898 & \textbf{0.006}  \\
  & hetero. & & \textbf{7.45} & \textbf{11.321} & \textbf{0.464}  & &  \textbf{8.73} & \textbf{10.739} & 0.044  \\
\cmidrule{2-10}
\multirow{2}*{{dPF-M}}
  & const.  & & 8.48 & 15.232 & 1.072  & &  \textbf{9.61} & 20.789 & \textbf{0.280}  \\
  & hetero. & & \textbf{8.23} & \textbf{14.725} & \textbf{0.787}  & &  9.76 & \textbf{19.833} & 0.413  \\
\bottomrule
\end{tabular}
\end{table*}

The effect of learning heteroscedastic process noise has not yet been evaluated 
in related work. We create datasets with heteroscedastic ground truth process 
noise, where the magnitude of $\mathbf{q}_v$ increases in three steps the closer
to the origin the disc is. The positional process noise  $\mathbf{q}_p$ remains 
constant ($\sigma_{q_p}=3.0$).

We compare the performance of DFs that learn constant and heteroscedastic 
process noise while he observation noise is heteroscedastic in all cases.

\paragraph{Results}

As shown in Table~\ref{tab:noise_qh}, learning heteroscedastic models 
of the process noise is a bit more difficult than for the observation noise. 
This is not surprising, as the input values for predicting the process noise are
the noisy state estimates.

Plotting the predicted values for $\mathbf{Q}$ (see Fig.~\ref{fig:hetero-q} 
for an example from the dEKF) reveals that all DFs learn to follow the real 
values for the heteroscedastic velocity noise relatively well, but also predict 
state dependent values for $\mathbf{q}_p$, which is actually constant. This 
could mean that the models have difficulties distinguishing between 
$\mathbf{q}_p$ and $\mathbf{q}_v$ as sources of uncertainty about the disc 
position. However, we see the same behavior also on a dataset with constant 
ground truth process noise. We thus assume that the models rather pick up an 
unintentional pattern in our data: The probability of the disc being occluded 
turned out to be higher in the middle of the image. The filters react to this by
overestimating $\mathbf{q}_p$ in the center, which results in an overall higher 
uncertainty about the state in regions where occlusions are more likely.

Despite not being completely accurate, learning heteroscedastic noise models 
still increases performance of all DFs by a small but reliable value. Even when 
the ground-truth process noise model is constant, most of the DFs were able to improve 
their RSME and likelihood scores slightly by learning ``wrong'' heteroscedastic 
noise models.

\subsubsection{Correlated Noise}

So far, we have only considered noise models with diagonal covariance matrices.
In this experiment, we want to see if DFs can learn to identify correlations 
in the noise. 
We compare the performance of DFs that learn noise models with diagonal or full 
covariance matrix on datasets with and without correlated process noise.
Both the learned process and the observation noise model are also heteroscedastic.

The results (see Appendix~\ref{app:sim-noise-corr}) show that 
learning correlated
noise models leads to a further small improvement of the performance 
of all DFs when the true process noise is correlated. However, uncovering 
correlations in the noise seems to be even more 
difficult than learning accurate heteroscedastic noise models, as indicated by
the still high Bhattacharyya distance between true and learned $\mathbf{Q}$.

\subsection{Benchmarking}

\begin{table}[ht!]
\centering
\caption{Results on disc tracking: Comparison between the DFs and LSTM 
models with one or two LSTM layers on two different datasets with 30 
distractors and constant process noise with 
increasing magnitude. Each experiment is repeated 
two times and we report mean and standard errors.
}\label{sim-tab-benchmark}
\footnotesize
{
\begin{tabular}{l  c c  c c} 
\toprule
&  \multicolumn{2}{c}{$\sigma_{q_p}=3.0$} &  \multicolumn{2}{c}{$\sigma_{q_p}=9.0$} \\
&  RMSE  & NLL                            &  RMSE  & NLL \\ 
\midrule
dEKF      &   6.31$\pm$0.12  &  9.24$\pm$0.10          &   11.83$\pm$0.28  & 11.10$\pm$0.20 \\
dUKF      &   6.46$\pm$0.20  &  9.26$\pm$0.26          &   11.49$\pm$0.18  & 10.75$\pm$0.16 \\
dMCUKF    &   6.53$\pm$0.18  &  \textbf{9.23$\pm$0.17} &   11.59$\pm$0.10  & \textbf{10.81$\pm$0.11} \\
dPF-M     &   6.75$\pm$0.07  & 12.33$\pm$0.09          &   11.52$\pm$0.07  & 20.50$\pm$0.36 \\
dPF-M-lrn &   \textbf{5.89$\pm$0.15}  & 11.43$\pm$0.15 &   \textbf{9.98$\pm$0.13}  & 19.17$\pm$0.18 \\
LSTM-1    & 	9.44$\pm$0.77  & 10.64$\pm$0.25        &   14.62$\pm$0.70  & 11.83$\pm$0.22 \\
LSTM-2    &   7.13$\pm$0.86  &  9.76$\pm$0.56          &   13.95$\pm$0.51  & 11.93$\pm$0.07 \\
\bottomrule
\end{tabular}
}
\end{table}

In the final experiment on this task, we compare the performance of the DFs among 
each other and to two LSTM models. We use an LSTM architecture similar to 
\cite{jonschkowski-2018}, with one or two layers of LSTM cells (512 units each). 
The LSTM state is decoded into mean and covariance of a Gaussian state estimate.

\paragraph{Experiment}
All models are trained for 30 epochs. The DFs learn the sensor and process 
models with heteroscedastic, diagonal noise models. We compare their performance on 
datasets with 30 distractors and different levels of constant or
heteroscedastic  process noise. Each experiment is repeated two times
to account for different initializations of the weights.

\paragraph{Results}

The results in Table~\ref{sim-tab-benchmark} show that all models (except
for the dPF-G, see Appendix Table~\ref{app:sim-tab-benchmark}) learn to track 
the target disc well and make reasonable uncertainty predictions. 
In terms of tracking error, the dPF with learned observation update 
performs best on all evaluated datasets. This, however, often does not extend to
the likelihood scores. For the NLL, the dMCUKF instead mostly achieves the best 
results, however, not with a significant advantage over the other DFs.

If we exclude the dPF variant with learned observation model (which is 
more expressive than the other DFs), 
we can see that the choice of the underlying filtering algorithm does not
make a big difference for the performance on this task. The unstructured
LSTM model, in contrast, requires two layers of LSTM cells (each with 512 
units per layer) to reach the performance of the DFs.
Unstructured models like LSTM can thus learn to perform similar to 
differentiable filters, but  require a much higher number of trainable 
parameters than the DFs which increases computational demands and the
risk of overfitting.

\section{KITTI Visual Odometry}\label{experiments-kitti}

\begin{table*}[ht!]
\centering
\caption{Results on \textit{KITTI-10}: Performance of the DFs with different 
noise models (mean and standard error).
Hand-tuned and Pretrained use fixed noise models whereas for the other variants, 
the noise models are trained end-to-end through the DFs. $\mathbf{R}_c$ indicates
a constant observation noise model and $\mathbf{R}_h$ a heteroscedastic one 
(same for $\mathbf{Q}$). The best results per DF are highlighted in bold.
}\label{tab-kitti-noise}
\footnotesize
{
\begin{tabularx}{0.9\linewidth}{c X Y Y Y Y Y Y Y} 
\toprule
& & \mbox{Hand-tuned} $\mathbf{R}_c\mathbf{Q}_c$ &  Pretrained $\mathbf{R}_c\mathbf{Q}_c$ & Pretrained $\mathbf{R}_h\mathbf{Q}_h$ & 
$\mathbf{R}_c\mathbf{Q}_c$ & $\mathbf{R}_c\mathbf{Q}_h$ & $\mathbf{R}_h\mathbf{Q}_c$ & $\mathbf{R}_h\mathbf{Q}_h$ \\
\midrule
\multirow{4}*{\rotatebox[origin=c]{90}{RMSE}}  
& dEKF   & 9.67$\pm$0.8  & \textbf{9.65$\pm$0.8}  & 10.53$\pm$1.0 & 9.70$\pm$0.8 & 9.69$\pm$0.8 & 9.74$\pm$0.8 & 9.68$\pm$0.8 \\
& dUKF   & 9.73$\pm$0.7  & \textbf{9.71$\pm$0.8}  & 10.68$\pm$1.0 & \textbf{9.71$\pm$0.8} & \textbf{9.71$\pm$0.8} & 9.81$\pm$0.8 & 9.72$\pm$0.8 \\
& dMCUKF & 9.73$\pm$0.7  & 9.71$\pm$0.8  & 10.68$\pm$1.0 & 9.71$\pm$0.8 & 9.70$\pm$0.8 & 9.80$\pm$0.8 & \textbf{9.68$\pm$0.8}  \\
& dPF-M  & 11.79$\pm$0.5 & 10.18$\pm$0.7 & 10.66$\pm$0.9 & \textbf{9.72$\pm$0.8} & 9.74$\pm$0.8 & 9.74$\pm$0.8 & 9.77$\pm$0.8 \\
\midrule
\multirow{4}*{\rotatebox[origin=c]{90}{NLL}} 
& dEKF   & 304.4$\pm$43.8 & 139.6$\pm$16.7 & 107.7$\pm$15.6    & 39.5$\pm$4.0 & 38.9$\pm$5.0          & 40.7$\pm$3.7 & \textbf{38.0$\pm$4.6} \\
& dUKF   & 305.9$\pm$43.7 & 140.0$\pm$16.6 & 108.1$\pm$15.5    & 40.5$\pm$4.0 & \textbf{39.2$\pm$5.1} & 41.3$\pm$4.0 & 40.1$\pm$5.4\\
& dMCUKF & 306.0$\pm$43.8 & 140.0$\pm$16.6 & 108.2$\pm$15.5    & 33.9$\pm$3.2 & \textbf{29.8$\pm$3.5} & 33.3$\pm$3.2 & 30.3$\pm$3.7 \\
& dPF-M  & 103.2$\pm$6.4  & 75.8$\pm$8.5   & \textbf{71.1$\pm$6.5} & 74.7$\pm$9.9 & 71.4$\pm$10.1     & 74.2$\pm$10.1 & 72.4$\pm$9.7 \\
\bottomrule
\end{tabularx}
}
\end{table*}

As a first real-world application we study the KITTI Visual Odometry problem 
\citep{geiger-2012} that was also evaluated by \cite{haarnoja-2016} and 
\cite{jonschkowski-2018}. 
The task is to estimate the position and orientation of a driving car given a 
sequence of RGB images from a front facing camera and the true initial state. 

The state is 5-dimensional and includes the position $\mathbf{p}$ and 
orientation $\theta$ of the car as well as the current linear and 
angular velocity $v$ and $\dot{\theta}$. The real control input 
$\mathbf{u} = \begin{pmatrix} \dot{v} & \ddot{\theta} \end{pmatrix}^T$ is unknown
and we thus treat changes in $v$ and $\dot{\theta}$ as results of the process
noise. The position and heading estimate can be updated analytically by Euler 
integration. 

While the dynamics model is simple, the challenge in this task comes from 
the unknown actions and the fact that the absolute position and orientation of 
the car cannot be observed from the RGB images. At each timestep, the filters 
receive the current images as well as a difference image between the current and
previous timestep. From this, the filters can estimate the angular and linear 
velocity to update the state, but the uncertainty about the absolute position and heading 
will inevitably grow due to missing feedback. 
Please refer to Appendix~\ref{app:kitti-imp} for details on
the implementation of the sensor network, the learned process model and the
learned noise models.

\subsection{Data}

The KITTI Visual Odometry dataset consists of eleven trajectories of varying length 
(from 270 to over 4500 steps) with ground truth annotations for position and 
heading and image sequences from two different cameras collected at 10\,Hz. 

Following \cite{haarnoja-2016} and \cite{jonschkowski-2018}, we build eleven
different datasets. Each of the original trajectories is used as
the test split of one dataset, while the remaining 10 sequences are used to 
construct the training and validation split.

To augment the data, we use the images from both cameras for each trajectory
and also mirror the sequences.
For training and validation, we extract 200 sequences of length 50 
with different random starting points from each augmented trajectory. This
results in 1013 training and 287 validation sequences. For 
testing, we extract sequences of length 100 from the augmented test-trajectory. 
The number of test sequences depends on the overall length of the test-
trajectory. 

When looking at the statistics of the eleven trajectories in the original KITTI
dataset, Trajectory 1 can be identified as an outlier: It shows driving on a 
highway, where the velocity of the car is much higher than in all the other 
trajectories. As a result, the sensor models trained on the other sequences will
yield bad results when evaluated on Trajectory 1. We will therefore mostly
report results for only a ten-fold cross-validation that excludes the dataset 
for testing on Trajectory 1. We will refer to this as \textit{KITTI-10} while the full, 
eleven-fold cross validation will be denoted as \textit{KITTI-11}. In 
Sec.~\ref{kitti-benchmarking}, results for both settings are reported, such
that the influence of Trajectory 1 becomes visible.

\begin{table*}[t!]
\centering
\caption{Results on \textit{KITTI-10}: RMSE and negative log likelihood for the DFs 
with different training schemes (mean and standard error). We compare 
individually trained process, sensor and noise models against finetuning only the 
sensor and process models, finetuning only the noise 
models and finteuning all models through the DFs. We also report results for DFs trained 
\textit{from scratch} without individual pretraining. The best results per DF 
are marked in bold.
}\label{tab-kitti-train}
\footnotesize
{
\begin{tabularx}{0.7\linewidth}{c X Y Y Y Y Y} 
\toprule
& & Individual &  Finetune Models & Finetune Noise & Finetune All & From Scratch \\
\midrule
\multirow{4}*{\rotatebox[origin=c]{90}{RMSE}}  
& dEKF   & 9.58$\pm$0.7  & 10.38$\pm$1.0 & \textbf{9.54$\pm$0.7}  & 9.83$\pm$0.8          & 10.05$\pm$0.8  \\
& dUKF   & 9.64$\pm$0.7  & 9.66$3\pm$0.8  & 9.57$\pm$0.7           & 9.33$\pm$0.8          & \textbf{9.29$\pm$0.6} \\
& dMCUKF & 9.64$\pm$0.7 &  9.53$\pm$0.8 & 9.58$\pm$0.7            & \textbf{9.35$\pm$0.7} & 9.72$\pm$0.6 \\
& dPF-M  & 10.29$\pm$0.6 & 10.86$\pm$0.8 & \textbf{9.59$\pm$0.6}  & 10.09$\pm$0.9          & 10.20$\pm$0.9 \\
\midrule
\multirow{4}*{\rotatebox[origin=c]{90}{NLL}} 
& dEKF   & 130.0$\pm$16.3 & 160.0$\pm$28.8 & \textbf{51.3$\pm$5.1} & 57.7$\pm$5.2     & 61.8$\pm$7.7 \\
& dUKF   & 126.7$\pm$15.6 & 118.4$\pm$14.5 & \textbf{57.3$\pm$5.5} & 87.1$\pm$9.9     & 59.3$\pm$7.2 \\
& dMCUKF & 127.9$\pm$15.9 & 117.8$\pm$14.8 & \textbf{50.0$\pm$4.6} & 74.4$\pm$8.1     & 50.3$\pm$8.1 \\
& dPF-M  &  76.9$\pm$7.6  &  86.3$\pm$11.3 & \textbf{72.6$\pm$9.4} & 80.9$\pm$12.2    & 82.4$\pm$12.2\\
\bottomrule
\end{tabularx}
}
\end{table*}

\subsection{Learning Noise Models}\label{experiments-kitti-noise}

In this experiment, we want to test how much the DFs profit from learning the
process and observation noise models end-to-end through the filters, as 
compared to using hand-tuned or individually learned noise models. 

We also again compare learning constant or
heteroscedastic noise models. In contrast to the previous task, we do not
expect as large a difference between constant or heteroscedastic observation
noise for this task, as the visual input does not contain occlusions or other 
events that would drastically change the quality of the predicted
observations $\mathbf{z}$. 

\paragraph{Experiment}

As in the experiments on simulated data (Sec.~\ref{sim-noise}), we use a 
fixed, pretrained sensor model and the analytical process model, and only train 
the noise models. We initialize $\mathbf{Q}$ and $\mathbf{R}$ with 
$\mathbf{Q} = \mathbf{I}_5$ and $\mathbf{R} = \mathbf{I}_2$. All DFs are 
trained with $L_{\mathrm{NLL}}$ and a sequence length of 25, which we found to be 
beneficial for learning the noise models in a preliminary experiment. 

We compare the DFs when learning different combinations of constant or 
heteroscedastic process and observation noise.
As on baseline, we use DFs with fixed constant noise models that reflect the average 
validation error of the pretrained sensor model and the analytical process model.
A second baseline fixes the noise models to those obtained by individual 
pretraining, where we evaluate both constant and heteroscedastic models.
All DFs are evaluated on \textit{KITTI-10}.

\paragraph{Results}

The results in Table~\ref{tab-kitti-noise} show that learning the noise 
models end-to-end through the filters greatly improves the NLL but has no big 
effect on the tracking errors for this task. The DFs with the hand-tuned, constant 
noise model have the by far worst NLL because they greatly 
underestimate the uncertainty about the vehicle pose. The DFs that use 
individually trained noise models perform better, but are still overly confident. 

For most of the DFs, we achieve the best results when learning constant 
observation and heteroscedastic process noise. The worst results are achieved when 
instead the observation noise is heteroscedastic and the process noise constant. This 
could indicate that the true process noise can be better modeled by a state-dependent
noise model while learning heteroscedastic observation noise leads to 
overfitting to the training data. However, the differences are overall not very 
pronounced.

Finally, we also evaluated the DFs with full covariance matrices for the noise 
models.
For the setting with constant observation and heteroscedastic process noise,
using full instead of diagonal covariance matrices barely had any effect on the 
tracking error and only slightly improved the NLL (e.g.\ from 27.1$\pm$5.0 to 
26.5$\pm$4.6 for the dEKF).

\subsection{End-to-End versus Individual Training}

Previous work \citep{jonschkowski-2018} has shown that end-to-end training 
through differentiable filters leads to better results than running the DFs with
models that were trained individually. Specifically, pretraining the models
individually and finetuning end-to-end resulted in the best tracking 
performance. As a possible explanation, the authors found that the individually 
trained process noise model predicted noise close to the ground truth whereas 
the end-to-end trained model overestimated to noise, which is believed to be 
beneficial for filter performance.  

Does this mean that end-to-end training through DFs mostly affects the noise 
models? To test this, we pretrain all models individually and compare the 
performance of the DFs without finetuning, when finetuning only the noise models
or only the sensor and process model and when finetuning everything. We also 
report results for training the DFs from scratch.

\paragraph{Experiment}

We pretrain sensor and process model and their associated (constant) noise 
models individually for 30 epochs. For finetuning, we load the pretrained models 
and finetune the desired parts for 10 epochs, while the end-to-end trained
versions are trained for 30 epochs. All variants are evaluated using 
\textit{KITTI-10} and trained using $L_{\mathrm{NLL}}$.

\paragraph{Results}

The results shown in Table~\ref{tab-kitti-train} support our 
hypothesis that end-to-end training through the DFs is most important for 
learning the noise models: Finetuning only the noise models improved both
RMSE and NLL of all DFs in comparison to the variants without finetuning or 
with finetuning only the sensor and process model (except for the dMCUKF). 
For dEKF and dPF, finetuning the sensor and process model even decreased 
the performance on both measures.

In terms of tracking error, individual pretraining plus finetuning the noise 
models lead to the best results on dEKF and dPF, while dUKF and dMCUKF performed
slightly better when finetuning both sensor and process model and their
noise models (dMCUKF) or even learning both from scratch (dUKF). 
For the NLL, finetuning only the noise models lead to the best results for 
all DFs, followed in most cases by training from scratch.

To summarize, the results indicate that individual pretraining is helpful 
for learning the sensor and process models, but not for the noise models. 
End-to-end training through the DFs, on the other hand, again proved to be 
important for optimizing the noise models for the respective filtering 
algorithm but did not offer advantages for learning the sensor and process 
model. 

\begin{table*}[ht!]
\centering
\caption{Results on KITTI: Comparison between the DFs and LSTM (mean and standard error).
Numbers for prior work BKF*, LSTM* taken from \cite{haarnoja-2016} and DPF* taken 
from \cite{jonschkowski-2018}. BKF* and DPF* use a fixed analytical process 
model while our DFs learn both, sensor and process model. 
$\frac{\mathrm{m}}{\mathrm{m}}$ and $\frac{\deg}{\mathrm{m}}$ denote the translation and rotation error at the final
step of the sequence divided by the overall distance traveled.
}\label{kitti-tab-benchmark}
\footnotesize
\begin{tabularx}{0.65\linewidth}{c X Y Y Y Y} 
\toprule
& & RMSE  & NLL  & $\frac{\mathrm{m}}{\mathrm{m}}$ & $\frac{\deg}{\mathrm{m}}$  \\
\midrule
\multirow{9}*{\rotatebox[origin=c]{90}{\textit{KITTI-11}}}  
& dEKF      &  15.8$\pm$5.8          & 338.8$\pm$277.1         & 0.24$\pm$0.04          & 0.080$\pm$0.005 \\
& dUKF      &  14.9$\pm$5.7          & 326.7$\pm$267.5         & $\mathbf{0.21\pm0.04}$ & 0.079$\pm$0.008 \\
& dMCUKF    &  15.2$\pm$5.5          & 266.3$\pm$216.1         & 0.23$\pm$0.04          & 0.083$\pm$0.012 \\
& dPF-M     &  16.3$\pm$6.1          & 115.2$\pm$34.6          & 0.24$\pm$0.04          & $\mathbf{0.078\pm0.006}$ \\
& dPF-M-lrn &  $\mathbf{14.3\pm5.2}$ & $\mathbf{94.2\pm33.3}$  & 0.22$\pm$0.04          & 0.088$\pm$0.013 \\
& LSTM      &  25.7$\pm$5.7          & 3970.6$\pm$2227.4       & 0.55$\pm$0.05          & 0.081$\pm$0.008 \\
\cmidrule{2-6}
& LSTM*& - & - & 0.26 & 0.29 \\
& BKF* & - & - & 0.21 & 0.08 \\
& DPF* & - & - & 0.15$\pm$0.015 & 0.06$\pm$0.009 \\
\midrule
\multirow{6}*{\rotatebox[origin=c]{90}{\textit{KITTI-10}}}  
& dEKF      &  10.1$\pm$0.8          & 61.8$\pm$7.7          & 0.21$\pm$0.03          & 0.079$\pm$0.006 \\
& dUKF      &  9.3$\pm$0.6           & 59.3$\pm$7.2          & $\mathbf{0.18\pm0.02}$ & 0.080$\pm$0.008 \\
& dMCUKF    &  9.7$\pm$0.6           & $\mathbf{50.3\pm8.1}$ & 0.2 $\pm$0.03          & 0.082$\pm$0.013 \\
& dPF-M     &  10.2$\pm$0.9          & 82.4$\pm$12.2         & 0.21$\pm$0.02          & $\mathbf{0.077\pm0.007}$ \\
& dPF-M-lrn &  $\mathbf{9.2\pm0.7}$  & 61.3$\pm$6.1          & 0.19$\pm$0.03          & 0.090$\pm$0.014 \\
& LSTM      &  20.2$\pm$2.0          & 1764.6$\pm$340.4      & 0.54$\pm$0.06          & 0.079$\pm$0.008 \\
\bottomrule
\end{tabularx}
\end{table*}

\subsection{Benchmarking}\label{kitti-benchmarking}

In the final experiment on this task, we compare the performance of the DFs 
to an LSTM model. We again use an LSTM architecture similar to 
\cite{jonschkowski-2018}, but with only one layer of LSTM cells with 256 units. 
The LSTM state is decoded into an update for the mean and the covariance 
of a Gaussian state estimate. Like 
the process model of the DFs, the LSTM does not get the full initial state as 
input, but only those components that are necessary for computing a state
update (velocities and sine and cosine of the heading). We chose
this architecture in an attempt to make the learning task easier for the LSTM.

\paragraph{Experiment}

All models are trained for 30 epochs using $L_{\mathrm{NLL}}$, except for the LSTM,
for which $L_{\mathrm{mix}}$ lead to better results. The DFs learn the 
sensor and process models with constant noise models. We report their 
performance on \textit{KITTI-10} and \textit{KITTI-11}, for comparison with 
prior work.

\paragraph{Results}

The results in Table~\ref{kitti-tab-benchmark} show that by training all the 
models in the DFs from scratch, we can reach a performance that is competitive 
with prior work by \cite{haarnoja-2016}, despite not relying on an analytical 
process model. We were, however, not able to reach the very good performance of
the dPF reported by \cite{jonschkowski-2018}. A possible cause for this
could be that the normalization of the particles in the learned observation
update used by \cite{jonschkowski-2018} helps the method to better deal with 
the higher overall velocity in Trajectory 1 of the KITTI dataset.

In contrast to the DF, we were not able to train LSTM models that reached a 
good evaluation performance on this task, despite trying multiple different 
architectures and loss functions. Different from the experiments on the
simulation task, increasing the number of units per LSTM-layer or using multiple 
LSTM layers even decreased the performance here. To complement our results, 
we also report an LSTM result from \cite{haarnoja-2016} that does better on 
the position error but worse on the orientation error. While these findings do 
not mean that a better performance could not be reached with unstructured models 
given different architectures or training routines, it still shows
that the added structure of the filtering algorithms greatly facilitates 
learning in more complex problems. 

For this task, the dPF-M-lrn again achieves the overall best tracking result, 
closely followed by the dUKF which reaches the lowest 
normalized endpoint position error ($\frac{\mathrm{m}}{\mathrm{m}}$).
One reason for the comparably bad performance of the  dEKF could be that the 
dynamics of the Visual Odometry task 
are more strongly non-linear than in the previous experiments. Both UKF and PF 
can convey the uncertainty more faithfully in this case, which could lead to  
better overall results when training on $L_{\mathrm{NLL}}$. Given the relatively large 
standard errors, the differences between the DFs are, however, not significant. 

\section{Planar Pushing}\label{experiments-pushing}

In the KITTI Visual Odometry problem, the main challenges were the unknown 
actions and dealing with the inevitably increasing uncertainty about the vehicle 
pose. With planar pushing, our second real-robot experiment in contrast addresses 
a task with much more complex dynamics. Apart from having non-linear and 
discontinuous dynamics (when the pusher makes or breaks contact with the object), 
\cite{bauza-2017} also showed that the noise in the system can be best captured 
by a heteroscedastic noise model.

With 10 dimensions, the state representation we use is also much larger than in 
our previous experiments. 
$\mathbf{x}$ contains the 2D position $\mathbf{p}_o$ and orientation $\theta$ of 
the object, as well as the two friction-related parameters $l$ and $\alpha_m$.
In addition, we include the 2D contact point between pusher and object 
$\mathbf{r}$, the normal to the object's surface at the contact point $\mathbf{n}$ 
and a contact indicator $s$. 
The control input $\mathbf{u}$ contains the start position $\mathbf{p}_u$ and 
movement $\mathbf{v}_u$ of the pusher.

An additional challenge of this task is that $\mathbf{r}$ and $\mathbf{n}$ are
only properly defined and observable when the pusher is in contact with the 
object. We thus set the labels for $\mathbf{n}$ to zeros and 
$\mathbf{r} = \mathbf{p}_u$ for non-contact cases.

\paragraph{Dynamics}

We use an analytical model by \cite{lynch-1992} to predict
the linear and angular velocity of the object ($\mathbf{v}_o$, $\omega$) given
the previous state and the pusher motion $\mathbf{v}_{u}$. However, 
predicting the next $\mathbf{r}$, $\mathbf{n}$ and $s$ is not possible with
this model since this would require access to a representation of the object shape.

For $\mathbf{r}$, we thus use a simple heuristic that predicts the next contact point
as $\mathbf{r}_{t+1} = \mathbf{r}_t + \mathbf{v}_{u,t}$. $\mathbf{n}$ and $s$ are 
only updated when the angle between pusher movement and (inwards facing) normal 
is greater than 90$^{\circ}$. In this case, we assume that the pusher moves away 
from the object and set $s_{t+1}$ and $\mathbf{n}_{t+1}$ to zeros. 

\paragraph{Observations}

Our sensor network receives simulated RGBXYZ images as input and outputs the
pose of the object, the contact point and normal as well as whether the push will
be in contact with the object during the push or not. 

Apart from from the latent parameters $l$ and $\alpha_m$, the orientation of the 
object, $\theta$, is the only state component that cannot be observed directly. 
Estimating the orientation of an object from a single image would require a
predefined ``zero-orientation'' for each object, which is impractical. Instead, 
we train the sensor network to predict the orientation relative to the object 
pose in the initial image of each pushing sequence. 

\subsection{Data}
\begin{figure}
\centering
\includegraphics[width=0.32\linewidth]{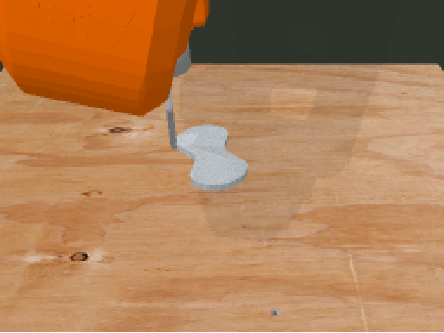}
\includegraphics[width=0.32\linewidth]{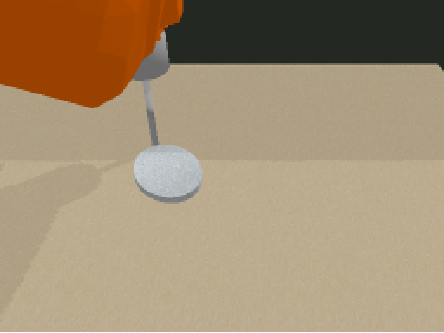}
\includegraphics[width=0.32\linewidth]{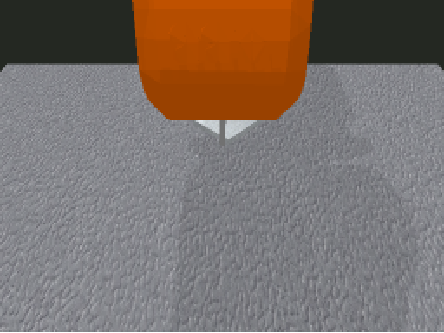}
\caption{Examples of the rendered RGB images that we use as observations
for the pushing task. The last example shows that the robot arm
can partially occlude the object in some positions.}\label{fig-pushing-data}
\end{figure}

We use the data from the MIT Push dataset \citep{yu-2016} as a basis for 
constructing our datasets. Further annotations for contact points and normals 
as well as rendered images are obtained using the tools described by 
\cite{kloss-2020}. 
However, in contrast to \cite{kloss-2020}, the images we use here also show the 
robot arm and are taken from a more realistic view-point. As a result, the 
robot frequently occludes parts of the object, but complete occlusions are rare. 
Figure~\ref{fig-pushing-data} shows example views.

We use pushes with a velocity of 50\,$\frac{\text{mm}}{\text{s}}$ and render 
images with a frequency of 5\,Hz. This results in short sequences of about five
images for each push in the original dataset. We extend them to 20 steps for 
training and validation and 50 steps for testing by chaining multiple pushes 
and adding in-between pusher movement when necessary. 
The resulting dataset contains 5515 sequences for training, 624 validation 
sequences and 751 sequences for testing.

\begin{table}[ht!]
\centering
\caption{Results for planar pushing:
Translation (tr) and rotation (rot) error and negative log likelihood for the
DFs with different noise models.
The hand-tuned DFs use fixed noise models whereas for the other variants, 
the noise models are trained end-to-end through the DFs. $\mathbf{R}_c$ indicates
a constant observation noise model and $\mathbf{R}_h$ a heteroscedastic one 
(same for $\mathbf{Q}$). The best result per DF are highlighted in bold.
}\label{tab-pushing-noise}
\footnotesize
{
\begin{tabularx}{\linewidth}{c l Y c c c c} 
\toprule
& & \mbox{Hand-tuned} $\mathbf{R}_c\mathbf{Q}_c$ &  $\mathbf{R}_c\mathbf{Q}_c$ &  $\mathbf{R}_h\mathbf{Q}_c$ &  $\mathbf{R}_c\mathbf{Q}_h$ & $\mathbf{R}_h\mathbf{Q}_h$ \\
\midrule
\multirow{4}*{\rotatebox[origin=c]{90}{tr [mm]}}  
& dEKF      & 6.22  & 4.45 & 4.61 & 4.44 & $\mathbf{4.38}$ \\
& dUKF      & 4.87  & 4.44 & 5.25 & $\mathbf{4.43}$ & 4.45 \\
& dMCUKF    & 4.73  & 4.42 & 4.8  & 4.39 & $\mathbf{4.35}$ \\
& dPF-M     & 18.13 & 5.07 & 4.92 & 5.32 & $\mathbf{4.64}$ \\
\midrule
\multirow{4}*{\rotatebox[origin=c]{90}{rot [$^{\circ}$]}}  
& dEKF      & 10.49            & 10.00 & $\mathbf{9.71}$  & 10.15 & 9.97 \\
& dUKF      & 9.87             & 9.91  & $\mathbf{9.73}$  & 10.05 & 10.00 \\
& dMCUKF    & $\mathbf{9.78}$  & 9.95  & 9.93             & 10.04 & 9.85 \\
& dPF-M     & 16.18            & 10.18 & $\mathbf{9.92}$  & 10.39 & 10.06 \\
\midrule
\multirow{4}*{\rotatebox[origin=c]{90}{NLL}}  
& dEKF      & 265.17 & 126.69 & 33.09  & 79.24  & $\mathbf{26.48}$ \\
& dUKF      & 378.08 & 84.12  & 33.06  & 81.55  & $\mathbf{27.61}$ \\
& dMCUKF    & 130.22 & 78.53  & 30.43  & 64.12  & $\mathbf{30.1}$ \\
& dPF-M     & 353.25 & 128.15 & 104.40 & 103.21 & $\mathbf{82.46}$ \\
\bottomrule
\end{tabularx}
}
\end{table}

\subsection{Learning Noise Models}

In this experiment, we again evaluate how much the DFs profit from learning the
process and observation noise models end-to-end through the filters.
In contrast to the KITTI task, for pushing, we expect both heteroscedastic 
observation and process noise to be advantageous, since the visual observations 
feature at least partial occlusions and the dynamics of pushing have been 
previously shown to exhibit heterostochasticity \citep{bauza-2017}.

To test this hypothesis, we compare DFs that learn constant or heteroscedastic 
noise models to DFs with hand-tuned, constant noise models that reflect the 
average test error of the pretrained sensor model and the analytical process 
model. 

\paragraph{Experiment}
As in the corresponding experiments on the previous tasks (Sec.~\ref{sim-noise} 
and Sec.~\ref{experiments-kitti-noise}), we use a fixed, pretrained sensor 
model and the analytical process model, and only train the noise models. All 
DFs are trained for 15 epochs on $L_{\mathrm{NLL}}$.

\paragraph{Results}
The results shown in Table~\ref{tab-pushing-noise}
again demonstrate that learning the noise models end-to-end through the structure of 
the filtering algorithms is beneficial. With learned models, all DFs reach much
better likelihood scores than with the hand-tuned variants. For the dEKF and 
especially the dPF, the tracking performance also improves significantly.

Comparing the results between constant and heteroscedastic noise models also
confirms our hypothesis that for the pushing task, heteroscedastic noise models are 
beneficial for both observation and process noise. While all DFs reach the best NLL when 
both noise models are state-dependent, the effect on the tracking error is, however,
less clear.

For dEKF, dUKF and dMCUKF, learning a heteroscedastic observation noise model leads to 
a much bigger improvement of the NLL than learning heteroscedastic process noise. Similar 
to the simulated disc tracking task, the input dependent noise model allows the
DFs to better deal with occlusions in the observations, which again reflects in a
negative correlation between the number of visible object pixels and the predicted
positional observation noise.

\subsection{Benchmarking}

\begin{table*}
\centering
\caption{Results on pushing: Comparison between the DFs and LSTM. 
Process and sensor model are pretrained and get finetuned end-to-end. The DFs learn
heteroscedastic noise models. Each experiment 
is repeated three times and we report mean and standard errors.
}\label{push-tab-benchmark}
\footnotesize
\begin{tabularx}{0.6\linewidth}{X Y Y Y Y} 
\toprule
&  RMSE  & NLL  & tr [mm] & rot [$^{\circ}$]  \\
\midrule
 dEKF      &   14.9$\pm$0.46 & 33.9$\pm$3.86  & $\mathbf{3.5\pm0.02}$  & 8.8$\pm$0.22 \\
 dUKF      &   $\mathbf{13.7\pm0.15}$ & $\mathbf{31.1\pm1.90}$  & 3.7$\pm$0.06  & 8.8$\pm$0.14 \\
 dMCUKF    &   13.8$\pm$0.10 & 34.1$\pm$3.57  & 3.7$\pm$0.06  & $\mathbf{8.8\pm0.06}$ \\
 dPF-M     &   18.3$\pm$0.38 & 120.4$\pm$5.70 & 5.7$\pm$0.16  & 10.5$\pm$0.36 \\
 dPF-M-lrn &   29.0$\pm$0.73 & 486.0$\pm$3.27 & 12.0$\pm$0.78 & 18.9$\pm$0.04 \\
 LSTM      &   27.36$\pm$0.2 & 35.4$\pm$0.24  & 8.8$\pm$0.17  & 19.0$\pm$0.001  \\
\bottomrule
\end{tabularx}
\end{table*}

In the final experiment, we compare the performance of the DFs 
to an LSTM model on the pushing task. As before, we use a model with one
LSTM layer with 256 units. 
The LSTM state is decoded into an update for the mean and the covariance 
of a Gaussian state estimate. 

\paragraph{Experiment}

All models are trained for 30 epochs using $L_{\mathrm{mix}}$. As initial experiments showed 
that learning sensor and process model jointly from scratch is very difficult for this task
due to the more complex architectures, we pretrain both models.
The sensor and process models are finetuned through the DFs and they learn heteroscedastic 
noise models. The LSTM, too, uses the pretrained sensor model, but not the process model.

\paragraph{Results}

As shown in Table~\ref{push-tab-benchmark}, even with a learned process model, all DFs 
(except for the dPF-M-lrn) perform at least similar to their pendants in the previous experiment
where we used the analytical process model. dEKF, dUKF and dMCUKF even reach a higher tracking
performance than before. As noted by \cite{kloss-2020}, this can be explained by the quasi-static 
assumption of the analytical model being violated for push velocities above 20\,$\frac{\text{mm}}{\text{s}}$.

The LSTM model, again, does not reach the performance of the DFs. One disadvantage of the
LSTM here is that in contrast to the DFs, we cannot isolate and pretrain the process model.
In contrast to the previous tasks, the dPF variant with the learned likelihood
function, however, performs even worse than the LSTM for planar pushing. This is likely due to 
the complex sensor model and the high-dimensional state that make learning the 
observation likelihood much more challenging.

\section{Conclusions}
\label{conclusion}

Our experiments show that all evaluated DFs are well suited for learning 
both sensor and process model, and the associated noise models. For simpler
tasks like the simulated tracking task and the KITTI Visual Odometry problem, all of 
these models can be learned end-to-end. Only the pushing problem
with its large state and complex dynamics and sensor model requires pretraining to
achieve good results.

In comparison to unstructured LSTM models, the DFs generally use fewer weights and 
achieve better results, especially on complex tasks. 
While training better LSTM models might be possible for more experienced LSTM users, 
using the algorithmic structure of the filtering algorithms definitely facilitated 
the learning problem and thus made it much easier to reach good performance with 
the DFs.
In addition, the structure of DFs allows us to pretrain components such as the 
process model that are not explicitly accessible in LSTMs.

The direct comparison between DFs with different underlying filtering algorithms
showed no clear winner. Only the dPF with learned observation update performed 
notably better than the other variants on the simulated example task and was least
affected by the outlier-trajectory of the KITTI-task. This variant relaxes some
of the assumptions that the filtering algorithms encode by not relying on an 
explicit sensor or observation noise model. Its good performance thus shows that the 
priors enforced by the algorithm choice can also be harmful if they do not hold in 
practice, such as  the Gaussian noise assumption.

Our experiments suggest that for learning the sensor and process model, end-to-end
training through the filters is convenient, but provides no advantages over training
the models individually. End-to-end training, however, proved to be 
essential for optimizing the noise models for their respective filtering algorithm.
In contrast to end-to-end trained models, both hand-tuned and individually trained 
noise models did not result in optimal performance of the DFs. 
Training noise models through DFs also enables learning more complex noise models 
than the ones used in learning-free, hand-tuned filters. We demonstrate that noise 
models with full (instead of diagonal) covariance matrices and especially heteroscedastic 
noise model, can significantly improve the tracking accuracy and uncertainty estimates 
of DFs.

The main challenge in working with differentiable filters is keeping the training  
stable  and  finding  good  choices for the numerous hyper-parameters and 
implementation options of the filters. While we hope that this work provides some 
orientation about which parameters matter and how to set them, we still recommend 
using the dEKF for getting started with differentiable filters. It is not only the most 
simple of the DFs we evaluated, but it also proved to be relatively insensitive to 
sub-optimal initialization of the noise models and was the most numerically stable during 
training. On the other hand, for tasks with strongly non-linear dynamics, the dUKF, dMCUKF 
or dPF can, however, ultimately achieve a better tracking performance.

One interesting direction for future research that we have not attempted here is 
to optimize parameters of the filtering algorithms, such as the scaling 
parameters of the dUKF or the fixed covariance of 
the mixture model components in the dPF-M, by end-to-end training. It could also be 
interesting to implement DFs with other underlying filtering
algorithms. For example, the pushing task could potentially
be better handled by a Switching Kalman filter \citep{murphy-1998} that explicitly 
treats the contact state as a binary decision variable.
In addition, all of our DFs perform badly on the outlier trajectory of the KITTI 
dataset which features a much higher driving velocity than the other trajectories
we used for training the model. This shows that the ability to detect input values 
outside of the training distribution would be a valuable addition to current DFs. 
Finally, it would be interesting to compare learning in DFs to similar variational
methods such as the ones introduced by \cite{karl-2017, fraccaro-2017, anh-2018} or 
the model-free PF-RNNs introduced by \cite{ma-2020}. 

\bibliographystyle{IEEEtranN}
\bibliography{cite.bib}

\newpage
\appendices

\renewcommand\thefigure{S\arabic{figure}}
\renewcommand\thetable{S\arabic{table}}
\renewcommand\theequation{S\arabic{equation}}
\setcounter{table}{0}
\setcounter{figure}{0}
\setcounter{equation}{0}
\section{Technical Background}
\label{app:filtering}

In the following section, we briefly review the Bayesian filtering algorithms 
that we use as basis for our differentiable filters. 

\subsection{Kalman Filter}\label{app:filtering:KF}

The Kalman filter \citep{kalman-1960} is a closed-form solution to the filtering 
problem for systems with a linear process and observation model and Gaussian 
additive noise:
\begin{align}
f(\mathbf{x}_{t}, \mathbf{u}_{t}) &= \mathbf{A}\mathbf{x}_{t} + \mathbf{B}\mathbf{u}_{t} + \mathbf{q}_t & \mathbf{q}_t \sim N(0, \mathbf{Q}_t) \\
h(\mathbf{x}_{t}) &= \mathbf{H}\mathbf{x}_{t} + \mathbf{r}_t & \mathbf{r}_t \sim N(0, \mathbf{R}_t)
\end{align}

The belief about the state $\mathbf{x}$ is represented by the 
mean $\boldsymbol{\mu}$ and covariance matrix $\boldsymbol{\Sigma}$ of a normal 
distribution. At each timestep, the filter predicts $\hat{\boldsymbol{\mu}}_t$ 
and $\hat{\boldsymbol{\Sigma}}_t$ using the process model. 
The innovation $\mathbf{i}_t$ is the difference between the predicted and actual 
observation and is used to correct the prediction. The Kalman Gain $\mathbf{K}$ 
trades-off the process noise $\mathbf{Q}$ and the observation noise $\mathbf{R}$ 
to determine the magnitude of the update.

\vspace{0.25cm}
\noindent Prediction Step:
\begin{align}
\hat{\boldsymbol{\mu}}_t &= \mathbf{A} \boldsymbol{\mu}_{t-1} + \mathbf{B} \mathbf{u}_t \label{app:filtering:KF:eq:mu} \\
\hat{\boldsymbol{\Sigma}}_t &= \mathbf{A} \boldsymbol{\Sigma}_{t-1} \mathbf{A}^T + \mathbf{Q}_{t-1} \label{app:filtering:KF:eq:sigma}
\end{align}

\noindent Update Step:

\vspace{-0.15cm}\noindent
\begin{tabular}{ll}
\parbox[t]{3.8cm}{
\begin{align}
\mathbf{S}_t &= \mathbf{H} \hat{\boldsymbol{\Sigma}}_t \mathbf{H}^T + \mathbf{R}_t \label{app:filtering:KF:eq:s}\\
\mathbf{K}_t &= \hat{\boldsymbol{\Sigma}}_t \mathbf{H}^T\mathbf{S}_t^{-1}  \label{app:filtering:KF:eq:kalmangain} \\
\mathbf{i}_t &= \mathbf{z}_t - \mathbf{H} \hat{\boldsymbol{\mu}}_t  \label{app:filtering:KF:eq:innovation} 
 \end{align}
} &
\parbox[t]{4.1cm}{
\begin{align}
\boldsymbol{\mu}_t &= \hat{\boldsymbol{\mu}}_t + \mathbf{K}_t \mathbf{i}_t     \label{app:filtering:KF:eq:mu_update} \\
\boldsymbol{\Sigma}_t &= (\mathbf{I}_n - \mathbf{K}_t \mathbf{H} ) \hat{\boldsymbol{\Sigma}}_t\label{app:filtering:KF:eq:sigma_update}
\end{align}
}
\end{tabular}

\subsection{Extended Kalman Filter (EKF)}
\label{app:filtering:EKF}

The EKF \citep{sorenson-1985} extends the Kalman filter to systems with 
non-linear process and observation models. It replaces the linear models for 
predicting $\hat{\boldsymbol{\mu}}$ in Equation~\ref{app:filtering:KF:eq:mu} and 
the corresponding observations $\hat{\mathbf{z}}$ in 
Equation~\ref{app:filtering:KF:eq:innovation} with non-linear models $f(\cdot)$ and $h(\cdot)$. For 
predicting the state covariance $\boldsymbol{\Sigma}$ and computing the 
Kalman Gain $\mathbf{K}$, these non-linear models are linearized around the 
current mean of the belief. The Jacobians  $\mathbf{F}_{|\mu_t}$ and 
$\mathbf{H}_{|\mu_t}$ replace $\mathbf{A}$ and $\mathbf{H}$ in Equations 
\ref{app:filtering:KF:eq:sigma} - \ref{app:filtering:KF:eq:kalmangain} 
and \ref{app:filtering:KF:eq:sigma_update}. This first-order approximation 
can be problematic for systems with strong non-linearity, as it does not take 
the uncertainty about the mean into account \citep{merwe-2004}.

\subsection{Unscented Kalman Filter (UKF)}
\label{app:filtering:UKF}

The UKF \citep{julier-1997, merwe-2004} was proposed to address the 
aforementioned problem of the EKF. Its core idea, the \textit{Unscented 
Transform} \citep{julier-1997}, is to represent a Gaussian random variable that 
undergoes a non-linear transformation by a set of specifically chosen points in 
state space, the so called \textit{sigma points} $\boldsymbol{\chi} \in 
\boldsymbol{\Chi}$. 

\begin{align}
\lambda             &= \alpha^2(\kappa + n) - n \label{app:filtering:UKF:eq:lambda} \\
\boldsymbol{\chi}^0 &= \boldsymbol{\mu} \nonumber \\ 
\boldsymbol{\chi}^i &= \boldsymbol{\mu} \pm (\sqrt{(n + \lambda)\boldsymbol{\Sigma}})_i  & & \forall i \in \{1...n\}    \label{app:filtering:UKF:eq:sigma_points} \\
w_m^0               &= \frac{\lambda}{\lambda + n}  &  w_c^0   &= \frac{\lambda}{\lambda + n} + (1 - \alpha^2 + \beta) \nonumber \\
w_{m}^i             &= w_{c}^i = \frac{0.5}{\lambda + n}     & & \forall i \in \{1...2n\}  \label{app:filtering:UKF:eq:weights}
\end{align}

Here, $n$ is the number of dimensions of the state $\mathbf{x}$. Each sigma 
point $\boldsymbol{\chi}^i$ has two weights $w_m^i$ and $w_c^i$.
The parameters $\alpha$ and $\kappa$ control the spread of the sigma points and how strongly 
the original mean $\boldsymbol{\chi}^0$ is weighted in comparison to the other 
sigma points. $\beta = 2$ is recommended if the true distribution of the system is 
Gaussian.

The statistics of the transformed random variable can then be calculated from 
the transformed sigma points. For example, in the prediction step of the UKF, 
the non-linear transform is the process model (Eq.
\ref{app:filtering:eq:UKF:f}) and the new mean and covariance of the 
belief are computed in Equations \ref{app:filtering:eq:UKF:mu} and 
\ref{app:filtering:eq:UKF:sigma}.

\begin{align}
\hat{\boldsymbol{\Chi}}_t &= f(\boldsymbol{\Chi}_{t-1}, \mathbf{u}_t) \label{app:filtering:eq:UKF:f} \\
\hat{\boldsymbol{\mu}}_t &= \sum_i w_m^i \hat{\boldsymbol{\chi}}^i_t  \label{app:filtering:eq:UKF:mu} \\
\hat{\boldsymbol{\Sigma}}_t &= \sum_i w_c^i (\hat{\boldsymbol{\chi}}^i_t - \hat{\boldsymbol{\mu}}_t)(\hat{\boldsymbol{\chi}}^i_t - \hat{\boldsymbol{\mu}}_t) ^T + \mathbf{Q}_t  \label{app:filtering:eq:UKF:sigma}
\end{align}

In the observation update step, $\mathbf{S}$, $\mathbf{K}$ and $\mathbf{i}$ 
from Equations \ref{app:filtering:KF:eq:s}, \ref{app:filtering:KF:eq:kalmangain}
and \ref{app:filtering:KF:eq:innovation} are likewise replaced by the following:

\begin{align}
\hat{\mathbf{z}}_t &= \sum_i w_m^i h(\hat{\boldsymbol{\chi}}^i_t) \\
\mathbf{S}_t &= \sum_i w_c^i (h(\hat{\boldsymbol{\chi}}^i_t) - \hat{\mathbf{z}}_t)(h(\hat{\boldsymbol{\chi}}^i_t) - \hat{\mathbf{z}}_t)^T + \mathbf{R}_t \\
\mathbf{K}_t &= \sum_i w_c^i (\hat{\boldsymbol{\chi}}^i_t) - \hat{\boldsymbol{\mu}}_t)(h(\hat{\boldsymbol{\chi}}^i_t) - \hat{\mathbf{z}}_t)^T\mathbf{S}_t^{-1} \\
\mathbf{i}_t &= \mathbf{z}_t - \mathbf{H} \hat{\boldsymbol{\mu}}_t
\end{align}

In theory, the UKF conveys the nonlinear transformation of the covariance more
faithfully than the EKF and is thus better suited for strongly non-linear 
problems \citep{thrun-2005}. In contrast to the EKF, it also does not require 
computing the Jacobian of the process and observation models, which can be 
advantageous when those models are learned.

In practice, tuning the parameters of the UKF can, however, sometimes be 
challenging. If $\alpha^2(\kappa + n)$ is too big, the sigma points are
spread too far from the mean and the prediction uncertainty increases.
However, for  $0 < \alpha^2(\kappa + n) < n$, the sigma point 
$\boldsymbol{\chi}^0$, which represents the original mean, is weighted 
negatively. This not only seems counter-intuitive, but strongly negative 
$w^0$ can also negatively affect the numerical stability of the 
UKF\citep{wu-2006}, which sometimes causes divergence of the estimated mean. 
In addition, if $\frac{\kappa}{n + \kappa} < 0$, the 
estimated covariance matrix is not guaranteed to be positive semi definite any 
more. This problem can be solved by changing the way in which 
$\boldsymbol{\Sigma}$ is computed (see Appendix III in \cite{julier-2000}).

\subsection{Monte Carlo Unscented Kalman Filter (MCUKF)}
\label{app:filtering:MCUKF}

The UKF represents the belief over the state with as few sigma points as 
possible. However, finding the correct scaling parameters $\alpha$, $\kappa$ and 
$\beta$ can sometimes be difficult, especially if the state is high dimensional. 
Instead of relying on the Unscented Transform to calculate the mean and 
covariance of the next belief, we can also resort to Monte Carlo methods, as 
proposed by \cite{wuthrich-2016-robust}. 

In practice, this means replacing the carefully constructed sigma points and 
their weights in Equations~\ref{app:filtering:UKF:eq:sigma_points} and
\ref{app:filtering:UKF:eq:weights} with uniformly weighted samples from the current 
belief. The rest of the UKF algorithm remains the same, but more sampled pseudo 
sigma points are necessary to represent the distribution of the belief 
accurately.

\subsection{Particle Filter (PF)}
\label{app:filtering:PF}

In contrast to the different variants of the Kalman filter explained before, the 
Particle filter \citet{gordon-1993} does not assume a parametric representation 
of the belief distribution. Instead, it represents the belief with a set of 
weighted \textit{particles}. This allows the filter to track multiple hypotheses 
about the state at the same time and makes it a popular choice for tasks like 
localization or visual object tracking \citep{thrun-2005}.

An initial set of particles $\boldsymbol{\chi}_0^i \in \boldsymbol{\Chi}_0$ is 
drawn from the initial belief and initialized with uniform weights $\pi$. 
In the prediction step, new particles are generated by applying the process 
model to the old particle set and sampling additive process noise:  
\begin{equation}
\boldsymbol{\Chi}_t = f(\boldsymbol{\Chi}_{t-1}, \mathbf{u}_t,
\mathbf{q}_t) \label{app:filtering:eq:PF:prediction}
\end{equation}

In the observation update step, the weight $\pi_t^i$ of each particle 
$\boldsymbol{\chi}_t^i$ is updated using current observation $\mathbf{z}_t$
by 
\begin{align}
\pi_t^i &= \pi_{t-1}^i p(\mathbf{z}_t | \boldsymbol{\chi}_t^i) & \forall \boldsymbol{\chi}_t^i \in \boldsymbol{\Chi}_t \label{app:filtering:eq:PF:update}
\end{align}

A potential problem of the PF is particle deprivation: Over time, many 
particles will receive a very low likelihood 
$p(\mathbf{z}_t | \boldsymbol{\chi}_t^i)$, and eventually the state would be 
represented by too few particles with high weights. To prevent this, a new set 
of particles with uniform weights can be drawn (with replacement) from the old 
set according to the weights. This resampling step focuses the particle set on 
regions of high likelihood and is usually applied after each timestep.

\section{Extended Experiments: Simulated Disc Tracking}

In the following, we present additional information about the experiments we 
performed for evaluating the DFs. This includes detailed information about
the network architectures for each task, extended results and additional 
experiments.

\subsection{Network Architectures and Initialization}\label{app:sim}

The network architectures for the sensor model and heteroscedastic observation
noise model are shown in Table~\ref{app:tab:sim-sensor}. 
Tables~\ref{app:tab:sim-process}
and \ref{app:tab:sim-process-noise} show the architecture for the learned 
process model and the heteroscedastic process noise. We denote fully connected 
layers by \textit{fc} and convolutional layers by \textit{conv}. 

\begin{table}
    \centering
   \caption{Sensor model and heteroscedastic observation noise architecture.
    Both fully connected output layers (for $\mathbf{z}$ and $diag(\mathbf{R})$) get 
    fc 2's output as input.}
    \label{app:tab:sim-sensor}
    \footnotesize
    \begin{tabularx}{\linewidth}{l c Y Y c}
    \toprule
    Layer & Output Size & Kernel & Stride & Activation \\
    \midrule
    Input $\mathbf{D}$ & $100\times100\times3$ & - & - & -  \\
    conv 1 & $50\times50\times4$ & $9\times9$ & 2 & ReLU  \\
    conv 2 & $25\times25\times8$ & $9\times9$ & 2 & ReLU  \\
    fc 1 & $16$ & - & - & ReLU  \\
    fc 2 & $32$ & - & - & ReLU  \\
    \midrule
    $\mathbf{z}$  & $2$ &  - & - & -  \\
    $\mathrm{diag}(\mathbf{R})$  & $2$ &  - & - & -  \\
    \bottomrule
    \end{tabularx}
\end{table}

\begin{table}
    \centering
    \caption{Learned process model architecture}
    \footnotesize
    \label{app:tab:sim-process}
    \begin{tabularx}{0.8\linewidth}{X Y Y}
    \toprule
    Layer & Output Size & Activation \\
    \midrule
    Input $\mathbf{x}$ & $4$ & -  \\
    fc 1 & 32 & ReLU \\
    fc 2 & 64 & ReLU \\
    fc 3 & 64 & ReLU \\
    \midrule
    $\Delta\mathbf{x}$ (fc) & 4 & - \\
    \bottomrule
    \end{tabularx}
\end{table}
\begin{table}
    \centering
    \caption{Heteroscedastic process noise model architecture}
    \footnotesize
    \label{app:tab:sim-process-noise}
    \begin{tabularx}{0.8\linewidth}{X Y Y}
    \toprule
    Layer & Output Size & Activation \\
    \midrule
    Input $\mathbf{x}$ & $4$ & -  \\
    fc 1 & 32 & ReLU \\
    fc 2 & 32 & ReLU \\
    \midrule
    $\mathrm{diag}(\mathbf{Q})$ (fc) & 4 & - \\
    \bottomrule
    \end{tabularx}
\end{table}

For the initial belief, we use $\boldsymbol{\Sigma}_{\mathrm{init}}=25*\mathbf{I}_4$. 
When training from scratch, we initialize $\mathbf{Q}$ 
and $\mathbf{R}$ with $\mathbf{Q} = 100 * \mathbf{I}_4$ and 
$\mathbf{R} = 900 * \mathbf{I}_2$, reflecting the high uncertainty of the 
untrained models.

\subsection{Implementation and Parameters}\label{app:sim-imp}

All experiments for evaluating different design choices and filter-specific 
parameters are performed on a dataset with 15 distractors and constant
process noise ($\sigma_p = 0.1, \sigma_{v} = 2$). The filters are trained 
end-to-end on $L_{\mathrm{NLL}}$ and learn the sensor and process model as well as 
heteroscedastic observation and constant process noise models.
We repeat each experiment two times to account for different initializations
of the weights and report mean and standard errors.

\subsubsection{dUKF}\label{app:sim-ukf}

\paragraph{Experiment}
The original version of the UKF by \cite{julier-1997} uses a simple 
parameterization where $\alpha=1$ and $\beta=0$ are fixed and only $\kappa$ 
varies. The authors recommend
setting $\kappa = 3 -n$. $\alpha$ and $\beta$ are used in the later proposed 
\textit{scaled unscented transform} \citep{julier-2002}, for which 
\cite{merwe-2004} suggest setting $\kappa=0$, $\beta=2$ and $\alpha$ to a 
small positive value.

We evaluate the original, simple parameterization as well as the one
for the scaled transform.
For the first, we test training the dUKF with $\kappa$ values in $[-10, 10]$. 
In the second case, we evaluate $\alpha \in \{0.001, 0.1, 0.5\}$ but do not 
vary $\beta$, for which the value 2 is optimal when working with Gaussians.

\paragraph{Results}
As discussed in Section~\ref{sim-ukf} of the main document, the results show 
no significant differences between the different parameter 
settings or between using the original parameterization from \cite{julier-1997} 
and the scaled transform. Only for $\kappa < -n$, the training failed 
due to a non-invertible matrix in the calculation of the Kalman Gain.

\subsubsection{dMCUKF}

\begin{figure}
\centering
\begin{tikzpicture}
	\begin{groupplot}[
		group style={
            group size=1 by 2,
            horizontal sep=1.25cm,
            vertical sep=0.35cm},
        enlarge x limits=0.2,
        width=8cm,
        height=4cm,
        xtick={data},
        xlabel = {\# particles or sigma points},
        ylabel style={yshift=0.05cm},
        xlabel style={yshift=-0.1cm},
        title style={yshift=-0.15cm, font=\scriptsize},
        ybar=0.pt,
        symbolic x coords={5, 10, 50, 100, 500},
		/pgf/bar width=0.35cm,
		 grid style={line width=.1pt, draw=gray!10},
        major grid style={line width=.2pt,draw=gray!40},
        ]
	\nextgroupplot[ylabel={tracking RMSE}, ymin=0, ymax=20, xticklabel=\empty, xlabel=\empty]
	  \addplot[mpi-red!70!black, fill=mpi-red!65,
	  		   error bars/.cd, y dir=both, y explicit, error bar style={line width=0.9pt}
	  		   ] table
        [x=samples, y=mcukf-dist, y error=mcukf-dist-std] {img/sim/samples}; 
      \addplot[mpi-blue!70!black, fill=mpi-blue!65, 
               error bars/.cd, y dir=both, y explicit, error bar style={line width=0.9pt}
	  		   ] table
        [x=samples, y=pf-ana-dist, y error=pf-ana-dist-std] {img/sim/samples};

    \nextgroupplot[ylabel={-log likelihood}, 
    				legend style={at={(0.45,-0.35)}},
			        legend columns=-1, 
			        legend entries={dMCUKF, dPF-M},
			        ymin=0, ymax=30]
	  \addplot[mpi-red!70!black, fill=mpi-red!65,
	  		   error bars/.cd, y dir=both, y explicit, error bar style={line width=0.9pt}
	  		   ] table
        [x=samples, y=mcukf-like, y error=mcukf-like-std] {img/sim/samples}; 
      \addplot[mpi-blue!70!black, fill=mpi-blue!65, 
               error bars/.cd, y dir=both, y explicit, error bar style={line width=0.9pt}
	  		   ] table
        [x=samples, y=pf-ana-like, y error=pf-ana-like-std] {img/sim/samples};  
    \end{groupplot}
\end{tikzpicture}
\caption{Results on disc tracking: Tracking error and negative log likelihood of the 
dMCUKF and dPF-M for different numbers of sampled sigma points or particles during 
training and 500 sigma points / particles for testing.}\label{app:fig-sim-samples}
\end{figure}
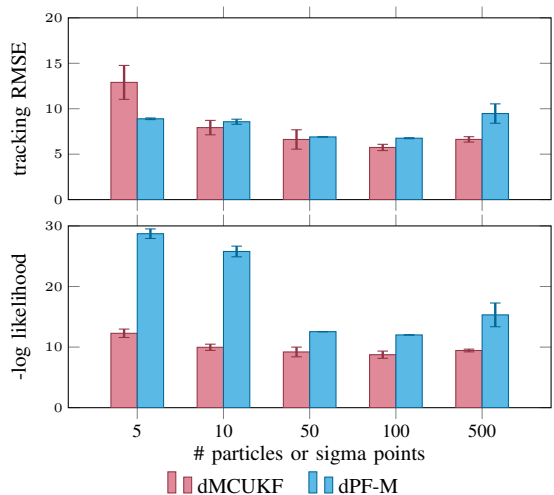

The results discussed in Section~\ref{sim-mcukf} of the main document 
are visualized in Figure~\ref{app:fig-sim-samples}.

\subsubsection{dPF: Belief Representation}\label{app:experiments-sim-pf-bel}

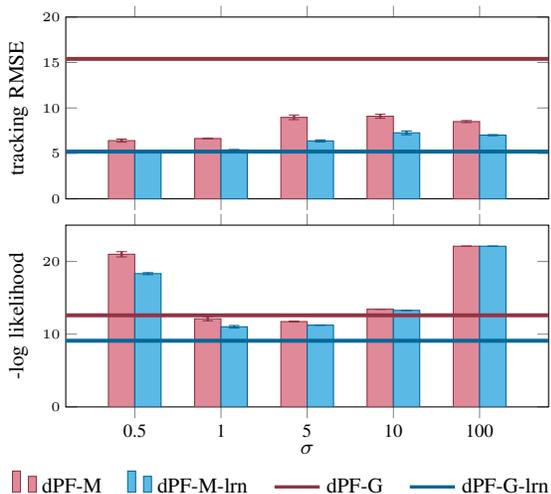
\begin{figure}
\centering
\begin{tikzpicture}
	\begin{groupplot}[
		group style={
            group size=1 by 2,
            vertical sep=0.35cm},
        enlarge x limits=0.2,
        width=8cm,
        height=4cm,
        xtick={data},
        ylabel style={yshift=0.05cm},
        xlabel style={yshift=-0.1cm},
        ybar=0.pt,
        symbolic x coords={0.1, 0.5, 1, 5, 10, 100},
		/pgf/bar width=0.35cm,
		 grid style={line width=.1pt, draw=gray!10},
        major grid style={line width=.2pt,draw=gray!40},
        ]
	\nextgroupplot[ylabel={tracking RMSE}, xticklabel=\empty, 
    			   ymin=0, ymax=20]
	  \addplot[mpi-red!70!black, fill=mpi-red!65,
	  		   error bars/.cd, y dir=both, y explicit, error bar style={line width=0.9pt}
	  		   ] table
        [x=mixture_std, y=ana_dist, y error=ana_dist_std] 
        {img/sim/pf_learning}; 
      \addplot[mpi-blue!70!black, fill=mpi-blue!65, 
	  		   error bars/.cd, y dir=both, y explicit, error bar style={line width=0.9pt}
	  		   ] table
        [x=mixture_std, y=lrn_dist, y error=lrn_dist_std] {img/sim/pf_learning}; 
      \draw[mpi-red!70!black, line width=1.5] ({rel axis cs:0,0}|-{axis cs:0.1,15.4}) --
                    ({rel axis cs:1,0}|-{axis cs:100,15.4});  
      \draw[mpi-blue!70!black, line width=1.5] ({rel axis cs:0,0}|-{axis cs:0.1,5.2}) --
                    ({rel axis cs:1,0}|-{axis cs:100,5.2});
    
    \nextgroupplot[ylabel={-log likelihood}, 
 				   xlabel = {$\sigma$},
    			   legend style={at={(0.45,-0.35)}},
			        legend columns=-1, 
			        legend entries={dPF-M, dPF-M-lrn, dPF-G, dPF-G-lrn},
			        ymin=0, ymax=25]
	  \addplot[mpi-red!70!black, fill=mpi-red!65, 
	  		   error bars/.cd, y dir=both, y explicit, error bar style={line width=0.9pt}
	  		   ] table
        [x=mixture_std, y=ana_like, y error=ana_like_std] {img/sim/pf_learning}; 
      \addplot[mpi-blue!70!black, fill=mpi-blue!65, 
	  		   error bars/.cd, y dir=both, y explicit, error bar style={line width=0.9pt}
	  		   ] table
        [x=mixture_std, y=lrn_like, y error=lrn_like_std] {img/sim/pf_learning}; 
      \draw[mpi-red!70!black, line width=1.5] ({rel axis cs:0,0}|-{axis cs:0.1,12.6}) --
                    ({rel axis cs:1,0}|-{axis cs:100,12.6});         
      \addlegendimage{line legend, sharp plot, mpi-red!70!black, line width=1.25}
      \draw[mpi-blue!70!black, line width=1.5] ({rel axis cs:0,0}|-{axis cs:0.1,9.1}) --
                    ({rel axis cs:1,0}|-{axis cs:100,9.1}); 
      \addlegendimage{line legend, sharp plot, mpi-blue!70!black, line width=1.25}
                    
    \end{groupplot}
\end{tikzpicture}
\caption{Results on disc tracking: Tracking error and negative log likelihood of 
the dPF-M and dPF-G, each with
using the analytical or learned (-lrn) observation update. The dPF-M and dPF-M-lrn are 
also evaluated for different values of the fixed per-particle covariance matrix 
$\boldsymbol{\Sigma} = \sigma^2 \mathbf{I}$ in the GMM.}\label{app:fig-sim-pf-like}
\end{figure}

The results discussed in Section~\ref{experiments-sim-pf-bel} are visualized 
in Figure~\ref{app:fig-sim-pf-like}. 

\subsubsection{dPF: Observation Update}\label{app:experiments-sim-pf-obs}

The likelihood for the observation update step of the dPF
can be implemented with an analytical Gaussian likelihood function (dPF-(G/M))
or with a neural network (dPF-(G/M)-lrn) as in \cite{jonschkowski-2018} and 
\cite{karkus-2018}. 

\cite{jonschkowski-2018} predict the likelihood based on an encoding of the 
sensory data and the observable components of the (normalized) particle states. 
Our implementation, too, takes the 64-dimensional encoding of the raw 
observations (fc 3 in Table~\ref{app:tab:sim-sensor}) and the observable 
particle state components as input. However, we decide not to normalize the 
particles, since having prior knowledge about the mean and standard deviation of 
each state component in the dataset might give an unfair advantage to the method 
over other variants.

\paragraph{Results}

Results for comparing the learned and analytical observation update can be
found in Figure~\ref{app:fig-sim-pf-like}.
Using a learned instead of an analytical likelihood function for updating the 
particle weights improves the tracking error of the dPF-M from 10.3$\pm$0.1 to 
8.3$\pm$0.1 and the NLL from 29.6$\pm$0.2 to 28.7$\pm$0.1. For the dPF-G, the 
difference is even more dramatic, with an RMSE of 23.3$\pm$1.1 vs. 8.0$\pm$0.3 
and an NLL of 31.0$\pm$0.05 vs. 27.5$\pm$0.1.

\subsubsection{dPF: Resampling}\label{app:experiments-sim-pf-re}

\begin{figure*}
\centering
\begin{tikzpicture}
	\begin{groupplot}[
		group style={
            group size=2 by 2,
            horizontal sep=1.cm,
            vertical sep=0.35cm},
        enlarge x limits=0.2,
        width=8cm,
        height=4cm,
        xtick={data},
        xlabel = {$\alpha_{\mathrm{re}}$},
        ylabel style={yshift=0.05cm},
        ybar=0.pt,
        symbolic x coords={0, 0.05, 0.1, 0.25},
		/pgf/bar width=0.3cm
        ]
        
   \nextgroupplot[ylabel={tracking RMSE}, title=dPF-M, ymin=0, 
				   ymax=20, xticklabels=\empty, xlabel=\empty]
	  \addplot[mpi-red!70!black, fill=mpi-red!65,
         error bars/.cd, y dir=both, y explicit, error bar style={line width=0.9pt}] table
        [x=alpha, y=dist_1, y error=dist_1_std] {img/sim/pf_resampling_ana}; 
      \addplot[mpi-blue!70!black, fill=mpi-blue!65,
               error bars/.cd, y dir=both, y explicit, error bar style={line width=0.9pt}] table
        [x=alpha, y=dist_2, y error=dist_2_std] {img/sim/pf_resampling_ana}; 
      \addplot[mpi-orange!70!black, fill=mpi-orange!65,
         error bars/.cd, y dir=both, y explicit, error bar style={line width=0.9pt}] table
        [x=alpha, y=dist_5, y error=dist_5_std] {img/sim/pf_resampling_ana}; 
      \addplot[mpi-lgreen!70!black, fill=mpi-lgreen!65,
         error bars/.cd, y dir=both, y explicit, error bar style={line width=0.9pt}] table
        [x=alpha, y=dist_5, y error=dist_5_std] {img/sim/pf_resampling_ana};

	\nextgroupplot[ylabel=\empty, title=dPF-M-lrn, ymin=0, 
				    ymax=20, xticklabels=\empty, xlabel=\empty]
	  \addplot[mpi-red!70!black, fill=mpi-red!65,
			   error bars/.cd, y dir=both, y explicit, error bar style={line width=0.9pt}] table
        [x=alpha, y=dist_1, y error=dist_1_std] {img/sim/pf_resampling_lrn}; 
      \addplot[mpi-blue!70!black, fill=mpi-blue!65,
   		       error bars/.cd, y dir=both, y explicit, error bar style={line width=0.9pt}] table
        [x=alpha, y=dist_2, y error=dist_2_std] {img/sim/pf_resampling_lrn}; 
      \addplot[mpi-orange!70!black, fill=mpi-orange!65,
               error bars/.cd, y dir=both, y explicit, error bar style={line width=0.9pt}] table
        [x=alpha, y=dist_5, y error=dist_5_std] {img/sim/pf_resampling_lrn}; 
      \addplot[mpi-lgreen!70!black, fill=mpi-lgreen!65,
              error bars/.cd, y dir=both, y explicit, error bar style={line width=0.9pt}] table
        [x=alpha, y=dist_5, y error=dist_5_std] {img/sim/pf_resampling_lrn};

    \nextgroupplot[ylabel={-log likelihood}, ymin=0, ymax=30]
    	  \addplot[mpi-red!70!black, fill=mpi-red!65,
         error bars/.cd, y dir=both, y explicit, error bar style={line width=0.9pt}] table
        [x=alpha, y=like_1, y error=like_1_std] {img/sim/pf_resampling_ana}; 
      \addplot[mpi-blue!70!black, fill=mpi-blue!65,
         error bars/.cd, y dir=both, y explicit, error bar style={line width=0.9pt}] table
        [x=alpha, y=like_2, y error=like_2_std] {img/sim/pf_resampling_ana}; 
      \addplot[mpi-orange!70!black, fill=mpi-orange!65,
         error bars/.cd, y dir=both, y explicit, error bar style={line width=0.9pt}] table
        [x=alpha, y=like_5, y error=like_5_std] {img/sim/pf_resampling_ana}; 
      \addplot[mpi-lgreen!70!black, fill=mpi-lgreen!65,
         error bars/.cd, y dir=both, y explicit, error bar style={line width=0.9pt}] table
        [x=alpha, y=like_5, y error=like_5_std] {img/sim/pf_resampling_ana};
        
    \nextgroupplot[ylabel=\empty, 
        		   legend style={at={(-0.1,-0.35)}},
			       legend columns=-1, 
			       legend entries={every step, every 2nd step, 
			                       every 5th step, every 10th step},
			       ymin=0, ymax=30]
	  \addplot[mpi-red!70!black, fill=mpi-red!65,
         error bars/.cd, y dir=both, y explicit, error bar style={line width=0.9pt}] table
        [x=alpha, y=like_1, y error=like_1_std] {img/sim/pf_resampling_lrn}; 
      \addplot[mpi-blue!70!black, fill=mpi-blue!65,
         error bars/.cd, y dir=both, y explicit, error bar style={line width=0.9pt}] table
        [x=alpha, y=like_2, y error=like_2_std] {img/sim/pf_resampling_lrn}; 
      \addplot[mpi-orange!70!black, fill=mpi-orange!65,
         error bars/.cd, y dir=both, y explicit, error bar style={line width=0.9pt}] table
        [x=alpha, y=like_5, y error=like_5_std] {img/sim/pf_resampling_lrn}; 
      \addplot[mpi-lgreen!70!black, fill=mpi-lgreen!65,
         error bars/.cd, y dir=both, y explicit, error bar style={line width=0.9pt}] table
        [x=alpha, y=like_5, y error=like_5_std] {img/sim/pf_resampling_lrn}; 

    \end{groupplot}
\end{tikzpicture}
\caption{Results on disc tracking: Tracking error and negative log likelihood 
of the two dPF-M variants for different 
resampling rates and values of the soft resampling parameter $\alpha_{\mathrm{re}}$.}
\label{app:fig-sim-resampling}
\end{figure*}
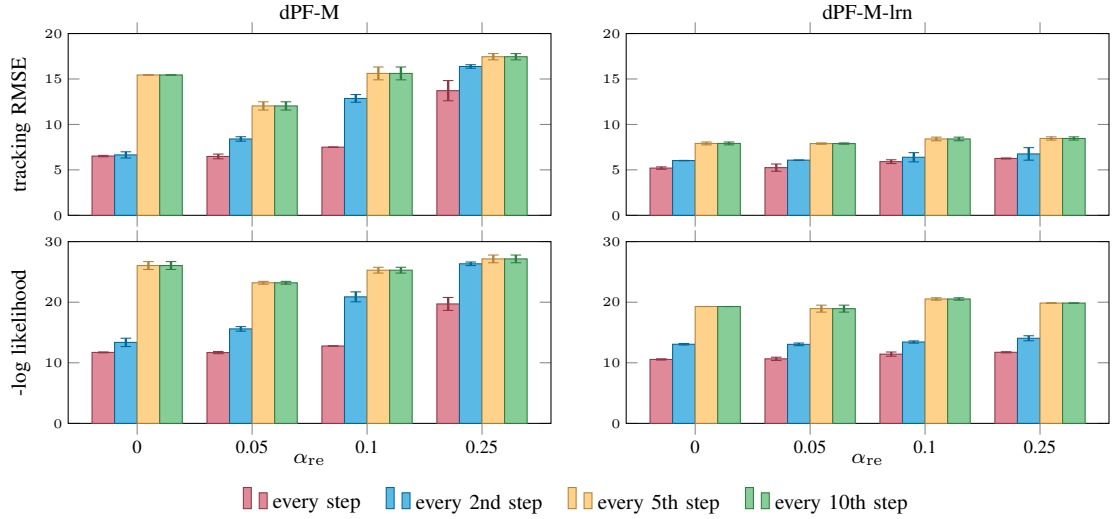

The results for Section~\ref{experiments-sim-pf-re} of the main document are 
visualized in Figure~\ref{app:fig-sim-resampling}.

\subsubsection{dPF: Number of Particles}

The results discussed in Section~\ref{experiments-sim-pf-num} of the main document 
are visualized in Figure~\ref{app:fig-sim-samples}.

\subsection{Noise Models}\label{app:sim-noise}

\subsubsection{Heteroscedastic Observation Noise}

Table~\ref{app:obs_noise} extends Table~\ref{tab:obs_noise} in the main
document. It contains results for the dPF-G and on additional datasets with 
different numbers of 
distractors and different magnitudes of the positional process noise.

\begin{table*}
\caption{Results for disc tracking: End-to-end learning of the noise models 
through the DFs on datasets
with 5 or 30 distractors and different levels of process noise. While 
$\mathbf{Q}$ is always constant, we evaluate learning constant (const.) or 
heteroscedastic (hetero) observation noise $\mathbf{R}$. We show the tracking 
error (RMSE), negative log likelihood (NLL), the correlation coefficient between 
predicted $\mathbf{R}$ and the number of visible pixels of the target disc 
(corr.) and the Bhattacharyya distance between true and learned process 
noise model ($D_{\mathbf{Q}}$). The best results per DF are highlighted in bold.
}\label{app:obs_noise}
\footnotesize
\centering
{
\begin{tabularx}{\linewidth}{Y l l  X   c c c c  X  c c c c  X  c c c c} 
\toprule
&  &   & & \multicolumn{4}{c}{$\sigma_{q_p}=0.1$}    & &  \multicolumn{4}{c}{$\sigma_{q_p}=3.0$}  & &  \multicolumn{4}{c}{$\sigma_{q_p}=9.0$} \\
&  & R & &  RMSE  & NLL   & corr. & $D_{\mathbf{Q}}$ & &  RMSE  & NLL  & corr. & $D_{\mathbf{Q}}$ & &  RMSE  & NLL   & corr. & $D_{\mathbf{Q}}$ \\
\midrule
\multirow{10}*{\rotatebox[origin=c]{90}{5 distractors}}  
& \multirow{2}*{{dEKF}} 
  & const.  & &   14.1 & 13.6 & -     & 2.722   & &  16.3 & 14.1 & -     & 0.081  & &  28.8 & 15.7 & -     & 0.019 \\
& & hetero. & &   \textbf{9.8}  & \textbf{11.9} & -0.71 & \textbf{1.204}  & &  \textbf{9.8}  & \textbf{11.5 }& -0.74 & \textbf{0.007}  & &  \textbf{18.7} & \textbf{13.2} & -0.66 & \textbf{0.007} \\
\cmidrule{2-18}
& \multirow{2}*{{dUKF}} 
  & const.  & &   14.3 & 13.7 & -     & 2.828  & &  17.1 & 14.2 & -     & 0.071  & &  30.2 & 15.8 & -     & 0.026 \\
& & hetero. & &   \textbf{9.9}  & \textbf{11.8} & -0.70 & \textbf{0.557}  & &  \textbf{9.6}  & \textbf{11.3} & -0.74 & \textbf{0.011}  & &  \textbf{21.7} & \textbf{14.2} & -0.66 & \textbf{0.013} \\
\cmidrule{2-18}
& \multirow{2}*{{dMCUKF}} 
  & const.  & &   14.5 & 13.7 & -     & 2.389  & &  16.5 & 14.2 & -     & 0.258  & &  30.7 & 15.8       & - & 0.02 \\
& & hetero. & &   \textbf{9.9}  & \textbf{11.8} & -0.71 & \textbf{0.272}  & &  \textbf{9.9}  & \textbf{11.6} & -0.73 & \textbf{0.016}  & &  \textbf{21.0} & \textbf{14.4} & -0.65 & \textbf{0.004} \\
\cmidrule{2-18}
& \multirow{2}*{{dPF-G}} 
& const.    & &   14.6 & 13.7 & - & \textbf{3.318} & & 17.3 & 14.1 & - & \textbf{0.257} & & 29.2 & 15.7 & - & \textbf{0.04} \\
& & hetero. & &   \textbf{12.0} & \textbf{12.8} & -0.47 & 3.348 & & \textbf{13.8} & \textbf{13.4} & -0.47 & 0.297 & & \textbf{23.2} & \textbf{14.6} & -0.63 & 0.064 \\
\cmidrule{2-18}
& \multirow{2}*{{dPF-M}} 
& const.    & &  13.1 & 34.7 & - & 3.408 & & 15.2 & 40.8 & - & \textbf{0.279} & & 27.7 & 52.9 & - & 0.745 \\
& & hetero. & &  \textbf{10.3} & \textbf{19.8} & -0.7 & \textbf{3.361} & & \textbf{11.3} & \textbf{23.1} & -0.67 & 0.424 & & \textbf{18.0} & \textbf{36.1} & -0.74 & \textbf{0.147} \\
\midrule
\multirow{10}*{\rotatebox[origin=c]{90}{30 distractors}}  
& \multirow{2}*{{dEKF}} 
  & const.  & &  14.5 & 13.9 & - & 2.543 & & 16.2 & 14.0 & - &  0.121 & & 28.6 & 15.6 & - &  0.010 \\
& & hetero. & &  \textbf{7.8} & \textbf{10.4} & -0.72 & \textbf{1.429} & & \textbf{8.8} & \textbf{10.7} & -0.78 & \textbf{0.002} & & \textbf{22.4} & \textbf{14.7} & -0.75 & \textbf{0.008} \\
\cmidrule{2-18}
& \multirow{2}*{{dUKF}} 
  & const.  & &  15.3 & 13.9 & - &  2.047 & & 16.8 & 14.1 & - &  0.161 & & 30.2 & 15.7 & -& 0.024 \\
& & hetero. & &  \textbf{7.8} & \textbf{10.4} & -0.71 & \textbf{1.565} & & \textbf{8.8} & \textbf{10.7} & -0.78 & \textbf{0.013} & & \textbf{20.6} & \textbf{14.8} & -0.85 & \textbf{0.010} \\
\cmidrule{2-18}
& \multirow{2}*{{dMCUKF}} 
  & const.  & &  14.8 & 13.9 & - & 2.955 & & 16.7 & 14.1 & - & 0.152 & & 29.8 & 15.7 & - & 0.022  \\
& & hetero. & &  \textbf{7.8} & \textbf{10.4} & -0.71 & \textbf{1.533} & & \textbf{9.0} & \textbf{10.9} & -0.78 & \textbf{0.006} & & \textbf{22.1} & \textbf{15.1} & -0.78 & \textbf{0.016}  \\
\cmidrule{2-18}
& \multirow{2}*{{dPF-G}} 
& const.    & &  15.2 & 13.8 & - & 3.433 & & 17.3 & 14.1 & - & \textbf{0.224} & & 29.1 & 15.6 & - & \textbf{0.047} \\
& & hetero. & &  1\textbf{0.4} & \textbf{11.9} & -0.71 & \textbf{3.103} & & \textbf{12.1} & \textbf{12.6} & -0.53 & 0.277 & & \textbf{20.8} & \textbf{14.4} & -0.86 & 0.090 \\
\cmidrule{2-18}
& \multirow{2}*{{dPF-M}} 
& const.    & &  14.1 & 33.6 & - & 3.396 & & 16.1 & 34.3 & - & 0.435 & & 27.7 & 49.9 & - & 1.240 \\
& & hetero. & &  \textbf{8.7} & \textbf{14.4} & -0.70 & \textbf{3.223} & & \textbf{9.6} & \textbf{20.8} & -0.77 & \textbf{0.280} & & \textbf{13.9} & \textbf{21.8} & -0.81 & \textbf{0.084} \\
\bottomrule
\end{tabularx}
}
\end{table*}

\subsubsection{Heteroscedastic Process Noise}

Table~\ref{app:tab:obs_noise_qh_full} extends Table~\ref{tab:noise_qh} in the main
document. It contains results for the dPF-G and on additional datasets with different 
magnitudes of the positional process noise.

\begin{table*}
\footnotesize
\centering
\caption{Results on disc tracking: End-to-end learning of constant or 
heteroscedastic process noise
$\mathbf{Q}$ on datasets with 30 distractors and different heteroscedastic
or constant ($\sigma_{q_p}=3.0$, $\sigma_{q_v}=2.0$) process noise. 
$D_{\mathbf{Q}}$ is the Bhattacharyya distance between true and learned process 
noise.
}\label{app:tab:obs_noise_qh_full}
\begin{tabular}{l l l c c c l  c c c  l c c c  l c c c} 
\toprule
  &   & & \multicolumn{3}{c}{hetero. $\sigma_{q_v}$, $\sigma_{q_p}=0.1$} & & \multicolumn{3}{c}{hetero. $\sigma_{q_v}$, $\sigma_{q_p}=3.0$}  &  & 
          \multicolumn{3}{c}{$\sigma_{q_p}=3.0$, $\sigma_{q_v}=2.0$}     & & \multicolumn{3}{c}{hetero. $\sigma_{q_v}$, $\sigma_{q_p}=9.0$} \\
  & Q & & RMSE  & NLL   &  $D_{\mathbf{Q}}$     & &  RMSE  & NLL   & $D_{\mathbf{Q}}$       & &  RMSE  & NLL    & $D_{\mathbf{Q}}$                           
      & & RMSE  & NLL    & $D_{\mathbf{Q}}$  \\
\midrule
\multirow{2}*{{dEKF}} 
  & const.  & &  4.3 & 8.2 & 1.335  & &  8.1 & 11.6 & 0.879  & &  8.8 & 10.7 & \textbf{0.002}  & &  19.6 & 14.1 & 1.108 \\
  &	hetero.	& & \textbf{3.8} & \textbf{7.2} & \textbf{0.479}  & &  \textbf{7.4} & \textbf{11.3 }& \textbf{0.402}  & &  8.8 & 10.7 & 0.033  & &  \textbf{18.5} & \textbf{13.7} & \textbf{0.805} \\
\midrule
\multirow{2}*{{dUKF}}
  & const.  & & 4.23 & 8.3 & 1.288  & &  7.8 & 11.3 & 0.874  & &  8.8 & 10.7 & \textbf{0.013}  & &  20.3 & 14.2 & 1.061 \\
  & hetero. & & \textbf{3.8} & \textbf{7.2} & \textbf{1.008}  & &  \textbf{7.6} & \textbf{11.2} & \textbf{0.391}  & &  \textbf{8.7} & 10.7 & 0.030 &  &  \textbf{20.1} & \textbf{14.0} & \textbf{0.900} \\
\midrule
\multirow{2}*{{dMCUKF}}
  & const.  & & 4.2 & 8.3 & 1.184  & &  8.1 & 11.5 & 0.891  & &  9.0 & 10.9 & \textbf{0.006}  & &  20.7 & 14.2 & 1.057 \\
  & hetero. & & \textbf{3.8} & \textbf{7.2} & \textbf{0.932} &  &  \textbf{7.5} & \textbf{11.3} & \textbf{0.464}  & &  \textbf{8.7} & \textbf{10.7} & 0.044  & &  \textbf{20.3} & \textbf{13.9} & \textbf{0.904} \\
\midrule
\multirow{2}*{{dPF-G}}
  & const.  & & 7.2 & 10.9 & \textbf{3.888}  & &  9.0 & 11.9 & 1.104  & &  12.1 & 12.6 & \textbf{0.277}  & &  \textbf{20.9} & 14.3 & 1.229 \\
  & hetero. & & \textbf{6.8} & \textbf{10.8} & 3.990  & &  9.0 & \textbf{11.5} & \textbf{0.808} & &  \textbf{11.8} & \textbf{12.4} & 0.347  & &  21.3 & 14.3 & \textbf{1.096} \\
\midrule
\multirow{2}*{{dPF-M}}
  & const.  & & \textbf{5.0} & \textbf{10.7} & 3.902  & &  8.5 & 15.2 & 1.072  & &  \textbf{9.6} & 20.8 & \textbf{0.280}  & &  19.7 & 29.1 & \textbf{0.799} \\
  & hetero. & & 5.2 & 11.1 & \textbf{3.853}  & &  \textbf{8.2} & \textbf{14.7} & \textbf{0.787}  & &  9.8 & \textbf{19.8} & 0.413  & &  \textbf{17.8} & \textbf{27.8} & 1.074 \\
\bottomrule
\end{tabular}
\end{table*}

\subsubsection{Correlated Noise}\label{app:sim-noise-corr}

So far, we have only considered noise models with diagonal covariance matrices.
In this experiment, we want to see if DFs can learn to identify correlations 
in the noise. 

\paragraph{Experiment}

We create a new dataset with 30 distractors and constant, correlated process noise. 
The ground truth process noise covariance matrix is
\begin{equation*}
\mathbf{Q}_{gt} = \begin{pmatrix}
 9. & -3.6 & 1.2 & 5.4 \\
-3.6 & 9. & -0.6 & 0. \\
 1.2 & -0.6 &  4. &  0. \\
 5.4 & 0. &  0. &  4. 
\end{pmatrix}
\end{equation*}

We compare the performance of DFs that learn noise models with diagonal or full 
covariance matrix on datasets with and without correlated process noise.
Both the learned process and the observation noise model are also heteroscedastic.

\paragraph{Results}

Results are shown in Table~\ref{app:tab-noise-corr}. Overall, we note that
learning correlated noise models has a small but consistent positive
effect on the tracking performance of all DFs, even when the ground truth
noise is not correlated. On the dataset with correlated ground truth 
process noise, we also observe an improvement of the likelihood scores. 

In terms of the Bhattacharyya distance between true and learned $\mathbf{Q}$, 
learning correlated models leads to a slight improvement for correlated ground 
truth noise and to slightly worse scores otherwise. This indicates that the models
are able to uncover some, but not all correlations in the underlying data.

In summary, while learning correlated noise models does not influence 
the results negatively, it also does not lead to a very pronounced improvement
over models with diagonal covariance matrices. Uncovering correlations in
the process noise thus seems to be even more difficult than learning 
accurate heteroscedastic noise models.

\begin{table*}
\footnotesize
\centering
\caption{Results on disc tracking: End-to-end learning 
of independent (\textit{diagonal} covariance 
matrix) or correlated (\textit{full} covariance matrix) process and 
observation noise models.
We evaluate on one dataset with independent, constant process noise  
($\sigma_{q_p}=3.0$, $\sigma_{q_v}=2.0$), one with independent heteroscedastic process
noise ($\sigma_{q_p}=3.0$), and one with correlated constant process noise.
$D_{\mathbf{Q}}$ is the Bhattacharyya distance between true and learned $\mathbf{Q}$.
}\label{app:tab-noise-corr}
{
\begin{tabularx}{0.9\linewidth}{l c l  Y Y Y l  Y Y Y  l Y Y Y } 
\toprule
  & covariance & & \multicolumn{3}{c}{independent const. noise} & & \multicolumn{3}{c}{independent hetero. noise} 
  & & \multicolumn{3}{c}{correlated const. noise}  \\
  & matrix     & & RMSE  & NLL   &  $D_{\mathbf{Q}}$ & & RMSE  & NLL   & $D_{\mathbf{Q}}$ & &
  RMSE  & NLL    & $D_{\mathbf{Q}}$   \\
\midrule
\multirow{2}*{{dEKF}} 
  & diagonal & & 8.8 & 10.7 & \textbf{0.033} & & \textbf{7.4} & \textbf{11.3} & \textbf{0.402} & & 8.9 & 10.6 & 1.249 \\
  &	full     & & \textbf{8.6} & \textbf{10.7} & 0.089 & & 7.6 & 12.2 & 0.591  & & \textbf{8.4} & \textbf{10.1} & \textbf{1.003} \\
\cmidrule{2-14}
\multirow{2}*{{dUKF}}
  & diagonal & & 8.7 & 10.7 & \textbf{0.030} & & \textbf{7.6} & 11.2 & \textbf{0.391} & & 8.7 & 10.6 & 1.345 \\
  & full     & & \textbf{8.6} & \textbf{10.7} & 0.126 & & 7.6 & \textbf{10.8} & 0.523 & & \textbf{8.7} & \textbf{10.4} & \textbf{0.994} \\
\cmidrule{2-14}
\multirow{2}*{{dMCUKF}}
  & diagonal & & 8.7 & 10.7 & \textbf{0.044} & & \textbf{7.5} & 11.3 & \textbf{0.464} & & 8.8 & 10.6 & 1.248 \\
  & full     & & \textbf{8.6} & \textbf{10.7} & 0.143 & & 7.6 & \textbf{10.8} & 0.507 & & \textbf{8.7} & \textbf{10.3} & \textbf{1.026}  \\
\cmidrule{2-14}
\multirow{2}*{{dPF-G}}
  & diagonal & & 11.8 & 12.4 & \textbf{0.347} & & 9.0 & 11.5 & \textbf{0.808} & & 11.7 & \textbf{12.4} & 1.646 \\
  & full     & & \textbf{11.5} & \textbf{12.3} & 0.421 & & \textbf{8.8} & 11.5 & 0.942 & & \textbf{11.4} & 12.5 & \textbf{1.565} \\
\cmidrule{2-14}
\multirow{2}*{{dPF-M}}
  & diagonal & & 9.8 & 19.8 & \textbf{0.413} & & 8.2 & \textbf{14.7} & \textbf{0.787} & & 9.3 & 22.1 & \textbf{1.649}  \\
  & full     & & \textbf{8.9} & \textbf{18.5} & 0.693 & & \textbf{7.3} & 15.7 & 1.463 & & \textbf{8.4} & \textbf{18.2} & 2.005 \\
\bottomrule
\end{tabularx}
}
\end{table*}

\subsection{Benchmarking}

Table~\ref{app:sim-tab-benchmark} extends Table~\ref{sim-tab-benchmark} from the 
main document. It contains results for the dPF-G and dPF-G-lrn and on additional 
datasets with lower positional process noise and heteroscedastic process noise.

\begin{table*}
\centering
\caption{Results on disc tracking: Comparison between the DFs and LSTM 
models with one or two LSTM layers on two different datasets with 30 
distractors and constant process noise with 
increasing magnitude. Each experiment is repeated 
two times and we report mean and standard error.
}\label{app:sim-tab-benchmark}
\footnotesize
{
\begin{tabularx}{0.95\linewidth}{l Y Y  Y Y  Y Y Y Y} 
\toprule
& \multicolumn{2}{c}{$\sigma_{q_p}=0.1$} &  \multicolumn{2}{c}{$\sigma_{q_p}=0.1$ hetero.} & 
 \multicolumn{2}{c}{$\sigma_{q_p}=3.0$} & \multicolumn{2}{c}{$\sigma_{q_p}=9.0$} \\
& RMSE  & NLL & RMSE  & NLL  & RMSE  & NLL   & RMSE  & NLL  \\
\midrule
dEKF      &  6.1$\pm$0.54  &  9.1$\pm$0.42           &   4.9$\pm$1.04  &  \textbf{8.3$\pm$0.91}  &   6.3$\pm$0.12  &  9.2$\pm$0.10          &  11.8$\pm$0.28  & 11.1$\pm$0.20 \\
dUKF      &  5.8$\pm$0.21  &  8.9$\pm$0.25           &   4.1$\pm$0.09  &  7.6$\pm$0.06           &   6.5$\pm$0.20  &  9.3$\pm$0.26          &  11.5$\pm$0.18  & 10.8$\pm$0.16 \\
dMCUKF    &  5.5$\pm$0.30  &  \textbf{8.6$\pm$0.23}  &   5.0$\pm$0.86  &  8.4$\pm$0.77           &   6.5$\pm$0.18  &  \textbf{9.2$\pm$0.17} &  11.6$\pm$0.10  & \textbf{10.8$\pm$0.11} \\
dPF-G     & 12.9$\pm$0.29  & 12.3$\pm$0.05           &  10.4$\pm$0.06  & 11.4$\pm$0.02           &  13.3$\pm$0.45  & 12.4$\pm$0.10          &  18.7$\pm$0.35  & 13.5$\pm$0.07 \\
dPF-M     &  6.2$\pm$0.34  & 11.7$\pm$0.16           &   5.0$\pm$0.40  & 10.7$\pm$0.45           &   6.7$\pm$0.07  & 12.3$\pm$0.09          &  11.5$\pm$0.07  & 20.5$\pm$0.36 \\
dPF-G-lrn &  \textbf{4.9$\pm$0.11}  &  9.2$\pm$0.12  &   \textbf{3.6$\pm$0.13}  &  8.4$\pm$0.04  &   \textbf{5.7$\pm$0.06}  &  9.8$\pm$0.01 &  10.6$\pm$0.17  & 11.9$\pm$0.03 \\
dPF-M-lrn &  5.3$\pm$0.17  & 10.8$\pm$0.22           &   4.4$\pm$0.09  &  9.9$\pm$0.10           &   5.9$\pm$0.15  & 11.4$\pm$0.15          &  \textbf{10.0$\pm$0.13}  & 19.2$\pm$0.18 \\
LSTM-1    &	  5.9$\pm$0.20  &  9.0$\pm$0.21           &  14.2$\pm$9.33  & 10.5$\pm$2.52           &   9.4$\pm$0.77  & 10.6$\pm$0.25          &  14.6$\pm$0.70  & 11.8$\pm$0.22 \\
LSTM-2    &  5.7$\pm$0.40  &  9.2$\pm$0.62           &   6.3$\pm$2.73  &  8.9$\pm$1.51           &   7.1$\pm$0.86  &  9.8$\pm$0.56          &  13.9$\pm$0.51  & 11.9$\pm$0.07 \\
\bottomrule
\end{tabularx}
}
\end{table*}

\section{Extended Experiments: KITTI Visual Odometry}\label{app:kitti}

\subsection{Network Architectures and Initialization}\label{app:kitti-imp} 

\paragraph{Sensor Network}

The network architectures for the sensor model and the heteroscedastic observation
noise model are shown in Table~\ref{app:tab:kitti-sensor}. At each timestep, 
the input consists of the current RGB image and the difference image between the
current and previous image.  The network architecture for the sensor model is 
the same as was used in \cite{haarnoja-2016} and  \cite{jonschkowski-2018}.

\begin{table*}
    \centering
    \caption{Sensor model and heteroscedastic observation noise architecture.
    Both output layers (for $\mathbf{z}$ and $\mathrm{diag}(\mathbf{R})$) get 
    fc 2's output as input.}
    \label{app:tab:kitti-sensor}
    \footnotesize
    \begin{tabularx}{0.8\linewidth}{X Y Y Y Y Y}
    \toprule
    Layer & Output Size & Kernel & Stride & Activation & Normalization\\
    \midrule
    Input $\mathbf{D}$ & $50\times150\times6$ & - & - & -  & - \\
    conv 1 & $50\times150\times16$ & $7\times7$ & $1\times1$ & ReLU & Layer \\
    conv 2 & $50\times75\times16$ & $5\times5$ & $1\times2$ & ReLU & Layer \\
    conv 3 & $50\times37\times16$ & $5\times5$ & $1\times2$ & ReLU & Layer \\
    conv 4 & $25\times18\times16$ & $5\times5$ & $2\times2$ & ReLU & Layer \\
    dropout (0.3) & $25\times18\times16$ & - & - & - & -  \\
    fc 1 & $128$ & - & - & ReLU & - \\
    fc 2 & $128$ & - & - & ReLU & - \\
    \midrule
    $\mathbf{z}$ (fc) & $2$ &  - & - & - & - \\
    $\mathrm{diag}(\mathbf{R})$ (fc) & $2$ &  - & - & -  & - \\
    \bottomrule
    \end{tabularx}
\end{table*}

\begin{table}
    \footnotesize
    \centering
    \caption{
    Learned process model architecture. We use a modified version of    
    the previous state $\mathbf{x}$ as input: 
    $\bar{\mathbf{x}} = ( v, \dot{\theta}, \cos \theta, \sin \theta )$ 
    }
    \label{app:tab:kitti-process}
    \begin{tabularx}{0.9\linewidth}{X Y Y}
    \toprule
    Layer & Output Size & Activation \\
    \midrule
    Input $\bar{\mathbf{x}}$ & $4$ & -  \\
    fc 1 & 32 & ReLU \\
    fc 2 & 64 & ReLU \\
    fc 3 & 64 & ReLU \\
    \midrule
    $\Delta\mathbf{x}$ (fc) & 5 & - \\
    \bottomrule
    \end{tabularx}
\end{table}
\begin{table}
    \centering
    \caption{Heteroscedastic process noise model architecture. We use a modified 
    version of  the previous state $\mathbf{x}$ as input: 
     $\bar{\mathbf{x}} = ( v, \dot{\theta}, \cos \theta, \sin \theta )$
    }
    \label{app:tab:kitti-process-noise}
    \begin{tabularx}{0.9\linewidth}{X Y Y}
    \toprule
    Layer & Output Size & Activation \\
    \midrule
    Input $\begin{pmatrix} v, \dot{\theta} \end{pmatrix}$ & $2$ & -  \\
    fc 1 & 32 & ReLU \\
    fc 2 & 32 & ReLU \\
    \midrule
    $\mathrm{diag}(\mathbf{Q})$ (fc) & 5 & - \\
    \bottomrule
    \end{tabularx}
\end{table}

\paragraph{Process Model}
Tables~\ref{app:tab:kitti-process} and \ref{app:tab:kitti-process-noise} show the 
architecture for the learned process model and the heteroscedastic process 
noise.
For both models, we found it to be important not to include the absolute 
position of the vehicle in the input values: The value range for the positions 
is not bounded, and especially for the dUKF variants, novel values encountered 
at test time often lead to a divergence of the filter. 

Excluding these values from the network inputs for predicting the state update 
also makes sense intuitively, since they are not required for computing the 
update analytically, either. For the state-dependent process noise, we not only 
exclude the position, but also the orientation of the car, as any relationships 
between vehicle pose and noise that could be learned would be specific to the 
training trajectories.

In addition, we provide the process model with the sine and cosine of $\theta$ 
as input instead of using the raw orientation, to facilitate the learning. In general,
dealing with angles in the state vector requires special attention: First, we 
correct angles to the range between $[-\pi, \pi]$ after every operation on the
state vector. Second, it is important to correctly calculate the difference
between angles (e.g.\ in the loss function) to avoid differences over 180$\deg$.
And third, computing the mean of several angles, e.g.\ for the particle mean
in the dPF, requires converting the angles to a vector representation.

\paragraph{Initialization}
When creating the noisy initial states, we do not add noise to the absolute 
position and orientation of the vehicle, since the DFs have no way of 
correcting them. We use 
$\mathrm{diag}(\boldsymbol{\Sigma}_{\mathrm{init}}) = \begin{pmatrix} 0.01 & 0.01 & 0.01 & 25 & 25
\end{pmatrix}$ for the initial covariance matrix. 
When training the DFs from scratch, we initialize the covariance matrices 
$\mathbf{Q}$ and $\mathbf{R}$ with 
$\mathrm{diag}(\mathbf{Q}) = \begin{pmatrix} 0.01 & 0.01 & 0.01 & 100 & 100
\end{pmatrix}$ and $\mathbf{R} = 100  \mathbf{I}_2$.
This reflects the high uncertainty of the untrained models, but also the fact 
that the process noise should be higher for the velocities (to account for
the unknown driver actions) than for the absolute pose.

\subsection{Training Sequence Length and Filter  Parameters}

One special feature of the Visual Odometry task is that the the error on the
estimated absolute vehicle pose will inevitably grow during filtering. As this 
could have an effect on the ideal training sequence length, we repeat the 
experiment from Section~\ref{experiments-sim-sl} in the main document.

For the dPF-M, we also evaluate different values of the fixed per-particle 
covariance $\boldsymbol{\Sigma}$ for calculating the GMM-likelihood. We 
anticipate that this parameter, too, could be sensitive to the accumulating 
uncertainty in the problem.

In addition, we also reevaluate different values for parameterizing the
sigma point selection and weighting in the dUKF.

\subsubsection{Training Sequence Length and dPF-M}

\paragraph{Experiment}

We only test with the dEKF, dUKF, dPF-M and dPF-M-lrn on \textit{KITTI-10}. 
The filters learn the sensor and process model as well as constant noise models. 
We train them using $L_{NLL}$ on sequence lengths 
$k \in \{2, 5, 10, 25\}$ while keeping the total number of examples per batch 
(steps $\times$ batch size) constant. 

For the dPF-M, we also evaluate two different values of the per-particle 
covariance, $\boldsymbol{\Sigma}= \mathbf{I}$ and 
$\boldsymbol{\Sigma}= 5^2 \mathbf{I}$. 

\paragraph{Results}

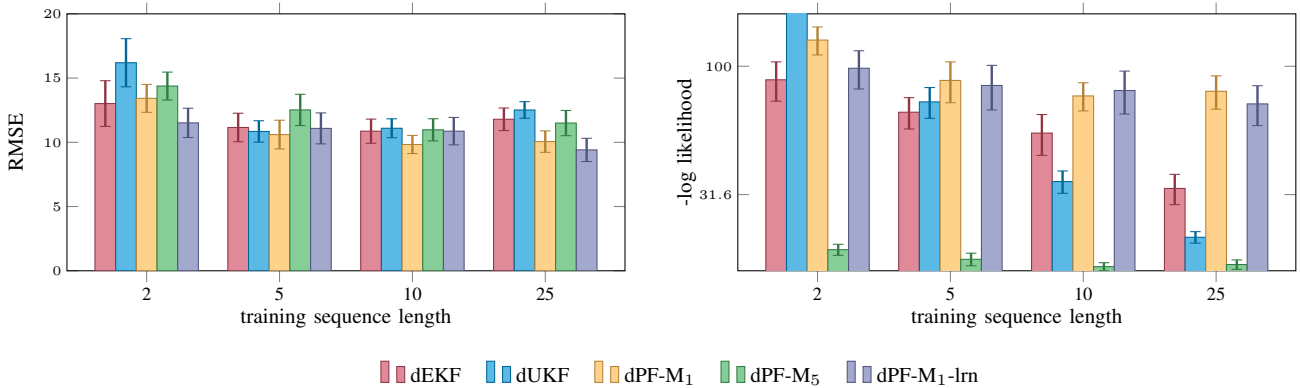
\begin{figure*}
\centering
\begin{tikzpicture}
	\begin{groupplot}[
		group style={
            group size=2 by 1, 
            horizontal sep=1.5cm,
            vertical sep=0.25cm},
        enlarge x limits=0.2,
        width=9cm,
        height=5cm,
        xtick={data},
        ylabel style={yshift=+0.1cm},
        xlabel style={yshift=-0.1cm},
        ybar=0.pt,
        symbolic x coords={2, 5, 10, 25}]
        
	\nextgroupplot[ylabel style={align=center}, ylabel={RMSE}, 
				   xlabel={training sequence length},  
				   ymin=0,				   
				   ymax=20,
				   bar width=0.275cm]
	  \addplot[mpi-red!70!black, fill=mpi-red!65, 
	  		   error bars/.cd, y dir=both, y explicit, 
	  		   error bar style={line width=0.9pt}] table
        [x=sl, y=ekf-dist, y error=ekf-dist-std] {img/kitti/kitti_sl};  
      \addplot[mpi-blue!70!black, fill=mpi-blue!65, 
	  		   error bars/.cd, y dir=both, y explicit, 
	  		   error bar style={line width=0.9pt}] table
        [x=sl, y=ukf-dist, y error=ukf-dist-std] {img/kitti/kitti_sl}; 
      \addplot[mpi-orange!70!black, fill=mpi-orange!65, 
	  		   error bars/.cd, y dir=both, y explicit, 
	  		   error bar style={line width=0.9pt}] table
        [x=sl, y=pf-m1-ana-dist, y error=pf-m1-ana-dist-std] {img/kitti/kitti_sl}; 
	\addplot[mpi-lgreen!70!black, fill=mpi-lgreen!65, 
	  		   error bars/.cd, y dir=both, y explicit, 
	  		   error bar style={line width=0.9pt}] table
        [x=sl, y=pf-m5-ana-dist, y error=pf-m5-ana-dist-std] {img/kitti/kitti_sl};    
    \addplot[mpi-purple!70!black, fill=mpi-purple!65, 
	  		   error bars/.cd, y dir=both, y explicit, 
	  		   error bar style={line width=0.9pt}] table
        [x=sl, y=pf-m1-lrn-dist, y error=pf-m1-lrn-dist-std] {img/kitti/kitti_sl};

    \nextgroupplot[ylabel style={align=center}, ylabel={-log likelihood}, 
    			   xlabel={training sequence length},  
    			   bar width=0.275cm, 
    			   ymax=160,
    			   ymin=0,
    			   ymode=log,
			       legend style={at={(-0.1,-0.35)}},
			       legend columns=-1, legend entries={dEKF, dUKF, dPF-M$_1$, dPF-M$_5$, dPF-M$_1$-lrn}]
	  \addplot[mpi-red!70!black, fill=mpi-red!65, 
	  		   error bars/.cd, y dir=both, y explicit, 
	  		   error bar style={line width=0.9pt}] table
        [x=sl, y=ekf-like, y error=ekf-like-std] {img/kitti/kitti_sl};  
      \addplot[mpi-blue!70!black, fill=mpi-blue!65, 
	  		   error bars/.cd, y dir=both, y explicit, 
	  		   error bar style={line width=0.9pt}] table
        [x=sl, y=ukf-like, y error=ukf-like-std] {img/kitti/kitti_sl}; 
      \addplot[mpi-orange!70!black, fill=mpi-orange!65, 
	  		   error bars/.cd, y dir=both, y explicit, 
	  		   error bar style={line width=0.9pt}] table
        [x=sl, y=pf-m1-ana-like, y error=pf-m1-ana-like-std] {img/kitti/kitti_sl}; 
	\addplot[mpi-lgreen!70!black, fill=mpi-lgreen!65, 
	  		   error bars/.cd, y dir=both, y explicit, 
	  		   error bar style={line width=0.9pt}] table
        [x=sl, y=pf-m5-ana-like, y error=pf-m5-ana-like-std] {img/kitti/kitti_sl};  
    \addplot[mpi-purple!70!black, fill=mpi-purple!65, 
	  		   error bars/.cd, y dir=both, y explicit, 
	  		   error bar style={line width=0.9pt}] table
        [x=sl, y=pf-m1-lrn-like, y error=pf-m1-lrn-like-std] {img/kitti/kitti_sl};  
        
    \end{groupplot}
\end{tikzpicture}
\caption{Results on \textit{KITTI-10}: Tracking error and negative log likelihood 
(NLL with logarithmic y axis) 
of dEKF, dUKF, dPF-M and dPF-M-lrn trained with different sequence lengths. 
For the dPF-M, we show two different values for the covariance $\boldsymbol{\Sigma}$ 
of the single Gaussians in the mixture model, $\boldsymbol{\Sigma}= \mathbf{I}$ 
and $\boldsymbol{\Sigma}= 5^2 \mathbf{I}$.
The cut-off NLL value for the dUKF on sequences of length 2 is 
2004.4$\pm$518.3.
}\label{app:fig-kitti-sl}
\end{figure*}

The results shown in Figure~\ref{app:fig-kitti-sl} largely confirm the results 
obtained for the simulation dataset in Section~\ref{experiments-sim-sl} of the main
document. 
We again see that longer training sequences increase the tracking performance of all
DFs up to a sequence length of around $k=10$. 

The dUKF seems to be most sensitive to the sequence length, with the highest 
tracking error and an extremely bad NLL score for sequences of length 2.
Different from the simulation experiment, for both dEKF and dUKF, the NLL 
keeps decreasing strongly over the full evaluated sequence length range,
despite the best RMSE already being reached at $k=5$. We attribute this 
to the accumulating uncertainty about the vehicle pose. For the dPFs, in 
contrast, the likelihood behaves similarly to the RMSE.

In light of the longer training times with higher sequence lengths, we again 
decide to keep a training-sequence length of 10 when training the DFs from 
scratch. However, when only the noise models are trained, longer sequences can 
be used to improved results on the NLL.

For the dPF-M, the experiment also shows that the covariance of the single 
distributions in the GMM is an important tuning parameter. With 
$\boldsymbol{\Sigma}= \mathbf{I}$,
we achieve the best tracking error, however, the likelihood does not reach the
performance of dEKF and dUKF. The NLL values can be drastically improved by 
using larger $\boldsymbol{\Sigma}$, at the cost of a decreased tracking
performance. Visual inspection of the position estimates shows that the particles 
remain relatively tightly clustered over the complete sequence, such that the 
likelihood of the GMM is not so different from the likelihood of the individual 
Gaussian components.  

This clustered particle distribution can be explained by the characteristics of 
the task: The uncertainty in the system mainly stems from the velocity components
that are affected by the unknown actions. However, by applying the observation update 
and resampling the particles at every step, we keep the variance in the velocity 
components small and thus prevent a stronger diffusion of the unobserved position 
components. This also explains why the dPF cannot profit as much as the dUKF and 
dEKF from seeing longer sequences during training. 

The large influence of the tuning parameter $\boldsymbol{\Sigma}$
on the value of the likelihood, independent of the tracking performance, also shows 
that comparing likelihood scores between different probabilistic models can be difficult.
In light of this, we decide to keep using $\boldsymbol{\Sigma}= \mathbf{I}$ for 
the better tracking error. 

\subsubsection{dUKF}

We also repeat the evaluation of different values of the parameters $\alpha$, $\kappa$ 
and $\beta$ for the dUKF described in Experiment~\ref{app:sim-ukf}. The experiment confirms 
our finding from the simulation experiment
that the exact choice of the values does not have a significant effect on the 
filter performance. We thus keep the values at $\alpha=1$, $\kappa=0.5$ and $\beta=0$.

\subsection{Learning Noise Models}\label{app:experiments-kitti-noise}

\paragraph{Experiment}

The baseline model with constant, hand-tuned noise uses \\
$\mathrm{diag}(\mathbf{Q}) = \begin{pmatrix}
10^{-4} & 10^{-4} & 10^{-6}&  0.01 & 0.16\end{pmatrix}^T$ and \\
$\mathrm{diag}(\mathbf{R}) = \begin{pmatrix} 0.36 & 0.36\end{pmatrix}^T$. 

\subsection{Benchmarking}\label{app:kitti-benchmarking}

Table~\ref{app:kitti-tab-benchmark} extends the results from 
Table~\ref{kitti-tab-benchmark} with data for the dPF-G and dPF-G-lrn.
Interestingly, we find that the difference in performance between the 
dPF variants with learned or analytical observation update is not as pronounced 
as in the results we obtained for the simulation experiment 
(Section~\ref{app:experiments-sim-pf-obs}). In particular, the dPF-G-lrn performs
similarly bad as the dPF-G on this task.

\begin{table*}[ht!]
\centering
\caption{Results on KITTI: Comparison between the DFs and LSTM (mean and standard error).
Numbers for prior work BKF*, LSTM* taken from \cite{haarnoja-2016} and DPF* taken 
from \cite{jonschkowski-2018}. BKF* and DPF* use a fixed analytical process 
model while our DFs learn both, sensor and process model. 
$\frac{\mathrm{m}}{\mathrm{m}}$ and $\frac{\deg}{\mathrm{m}}$ denote the translation and rotation error at the final
step of the sequence divided by the overall distance traveled.
}\label{app:kitti-tab-benchmark}
\footnotesize
\begin{tabularx}{0.75\linewidth}{c l Y Y Y Y} 
\toprule
& & RMSE  & NLL  & $\frac{\mathrm{m}}{\mathrm{m}}$ & $\frac{\deg}{\mathrm{m}}$  \\
\midrule
\multirow{11}*{\rotatebox[origin=c]{90}{\textit{KITTI-11}}}  
& dEKF      &  15.8$\pm$5.8          & 338.8$\pm$277.1         & 0.24$\pm$0.04          & 0.080$\pm$0.005 \\
& dUKF      &  14.9$\pm$5.7          & 326.7$\pm$267.5         & $\mathbf{0.21\pm0.04}$ & 0.079$\pm$0.008 \\
& dMCUKF    &  15.2$\pm$5.5          & 266.3$\pm$216.1         & 0.23$\pm$0.04          & 0.083$\pm$0.012 \\
& dPF-M     &  16.3$\pm$6.1          & 115.2$\pm$34.6          & 0.24$\pm$0.04          & $\mathbf{0.078\pm0.006}$ \\
& dPF-G     &  21.1$\pm$5.7         & 121.9$\pm$80.5          & 0.33$\pm$0.04          & 0.175$\pm$0.036 \\
& dPF-M-lrn &  $\mathbf{14.3\pm5.2}$ & $\mathbf{94.2\pm33.3}$  & 0.22$\pm$0.04          & 0.088$\pm$0.013 \\
& dPF-G-lrn &  19.1$\pm$5.3         & 197.8$\pm$125.3         & 0.31$\pm$0.06          & 0.168$\pm$0.049 \\
& LSTM      &  25.7$\pm$5.7          & 3970.6$\pm$2227.4       & 0.55$\pm$0.05          & 0.081$\pm$0.008 \\
\cmidrule{2-6}
& LSTM*& - & - & 0.26 & 0.29 \\
& BKF* & - & - & 0.21 & 0.08 \\
& DPF* & - & - & 0.15$\pm$0.015 & 0.06$\pm$0.009 \\
\midrule
\multirow{8}*{\rotatebox[origin=c]{90}{\textit{KITTI-10}}}  
& dEKF      &  10.1$\pm$0.8          & 61.8$\pm$7.7          & 0.21$\pm$0.03          & 0.079$\pm$0.006 \\
& dUKF      &  9.3$\pm$0.6           & 59.3$\pm$7.2          & $\mathbf{0.18\pm0.02}$ & 0.080$\pm$0.008 \\
& dMCUKF    &  9.7$\pm$0.6           & $\mathbf{50.3\pm8.1}$ & 0.2 $\pm$0.03          & 0.082$\pm$0.013 \\
& dPF-M     &  10.2$\pm$0.9          & 82.4$\pm$12.2         & 0.21$\pm$0.02          & $\mathbf{0.077\pm0.007}$ \\
& dPF-G     & 15.5$\pm$1.4          & 41.7$\pm$6.6          & 0.3 $\pm$0.04          & 0.182$\pm$0.038 \\
& dPF-M-lrn &  $\mathbf{9.2\pm0.7}$  & 61.3$\pm$6.1          & 0.19$\pm$0.03          & 0.090$\pm$0.014 \\
& dPF-G-lrn & 14.4$\pm$2.4          & 73.6$\pm$17.9         & 0.29$\pm$0.06          & 0.179$\pm$0.053 \\
& LSTM      &  20.2$\pm$2.0          & 1764.6$\pm$340.4      & 0.54$\pm$0.06          & 0.079$\pm$0.008 \\
\bottomrule
\end{tabularx}
\end{table*}

\section{Extended Experiments: Planar Pushing}\label{app:push}

\subsection{Network Architectures and Initialization}\label{app:push-archi}

\paragraph{Sensor Network}

Our architectures for the sensor network is very similar to the one used by 
\cite{kloss-2020}, where only the object position $\mathbf{p}_o$ is estimated 
from the full image while the contact-related state components ($\mathbf{r}$, 
$\mathbf{n}$, $s$) are computed from a smaller glimpse around the pusher location.

For predicting the orientation of the object, we extract a second glimpse from 
the full image, this time centered on the estimated object position. A small CNN 
then  predicts the change in orientation between the glimpse extracted
from the initial image in the sequence and the glimpse at the current time step.

\begin{figure*}
\centering
\includegraphics[width=0.8\linewidth]{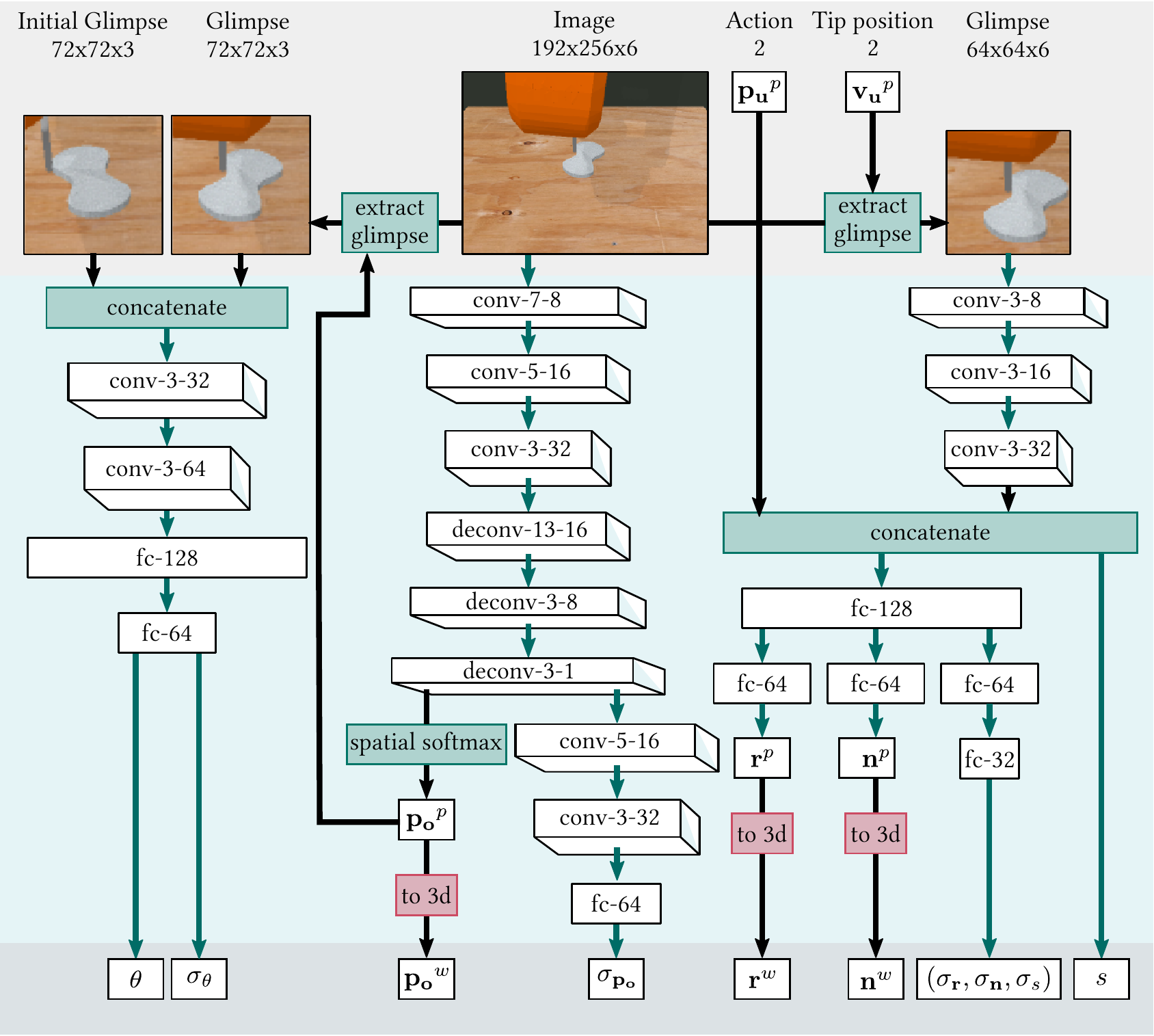}
\caption{
Architecture of the sensor network and heteroscedastic observation noise model for
planar pushing. We use 6-channel RGBXYZ images as input for computing the
object position and contact related state components. The object orientation is
estimated relative to the initial orientation by comparing the RGB glimpse 
centered on the current estimated object position to the initial one.
\\
White boxes represent tensors, green arrows and boxes indicate network layers, 
whereas black arrows represent dataflow without processing. For convolution (conv)
and deconvolution (deconv) layers, the numbers in each tensor are the kernel size
and number of output channels of the layer that produced it. For fully connected 
layers (fc), the number corresponds to  the number of output channels. \\
With the exception of the output layers, all convolution, deconvolution and 
fully connected layers are followed by ReLU non-linearities. The (de)convolution 
layers also use layer normalization.
}\label{app:fig-pushing-obs}
\end{figure*}

The sensor network predicts object position, contact point and normal in pixel
space because predictions in this space can be most directly related to the input 
image and the predicted feature maps. To this end, 
we also transform the action into pixel space before using it (together with the 
glimpse encoding) as input for predicting the contact point and normal. 
The pixel predictions are then transformed back to to world-coordinates using the 
depth measurements and camera information. 
The resulting sensor network including the layers for computing the heteroscedastic 
observation noise is illustrated in Figure~\ref{app:fig-pushing-obs}. 

\begin{table}
    \footnotesize
    \centering
    \caption{Learned process model architecture.}
    \label{app:tab:push-process}
    \begin{tabularx}{0.8\linewidth}{X Y Y}
    \toprule
    Layer & Output Size & Activation \\
    \midrule
    Input $(\mathbf{x}, \mathbf{v}_u)$ & 12 & -  \\
    fc 1 & 256 & ReLU \\
    fc 2 & 128 & ReLU \\
    fc 3 & 128 & ReLU \\
    \midrule
    $\Delta\mathbf{x}$ (fc) & 10 & - \\
    \bottomrule
    \end{tabularx}
\end{table}
\begin{table}
    \centering
    \caption{Heteroscedastic process noise model architecture. We use a 
    modified version of the previous state $\mathbf{x}$ as input: 
    $\bar{\mathbf{x}}$ does not include the latent parameter $l$.}
    \label{app:tab:push-process-noise}
    \begin{tabularx}{0.8\linewidth}{X Y Y}
    \toprule
    Layer & Output Size & Activation \\
    \midrule
    Input $(\bar{\mathbf{x}}, \mathbf{v}_u)$ & $11$ & -  \\
    fc 1 & 128 & ReLU \\
    fc 2 & 64 & ReLU \\
    \midrule
    $\mathrm{diag}(\mathbf{Q})$ (fc) & 10 & - \\
    \bottomrule
    \end{tabularx}
\end{table}

\paragraph{Process Model}

Tables~\ref{app:tab:push-process} and \ref{app:tab:push-process-noise} show the 
architecture for the learned process model and the heteroscedastic process 
noise. One problem we noticed is that the estimates for $l$ sometimes diverge during
filtering if the DFs estimate that the pusher is in contact with the object while it is not. 
Just as for the absolute position of the vehicle in the KITTI task, we thus found it
important for the stability of the dUKF and dMCUKF to not make the heteroscedastic
process noise model dependent on $l$.

Note that in the filter state, we measure $\mathbf{p}_o$ and $\mathbf{r}$ in 
millimeter and $\theta$ and $\alpha_m$ in degree. To avoid
having too large differences between the magnitudes of the state components,
we downscale $l$ by a factor of 100. 
$\mathbf{n}$ is a dimensionless unit vector and $s$ should take values between 
0 and 1.

To keep the filters stable during training, we found it necessary to enforce 
maximum and minimum values for $\alpha_m$ and $l$.
Both $\alpha_m$ and $l$ cannot become negative. The opening angle of the friction 
cone, $\alpha_m$, should also not be larger than $90^{\circ}$, while we limit $l$ to 
be in the range of $[0.1, 5000]$ to ensure that the computations in the 
analytical model remain numerically stable.

\paragraph{Initialization}
For the initial covariance matrix, we use 
\begin{align*}
&\sqrt{\mathrm{diag}(\boldsymbol{\Sigma}_{\mathrm{init}})} = \\
&\hspace{2em} \begin{pmatrix}
50 & 50 & 10^{-3} & 5 & 5 & 50 & 50 & 0.5 & 0.5 & 0.5
\end{pmatrix}^T
\end{align*}
When training the noise models, we initialize $\mathbf{Q}$ and $\mathbf{R}$ with 
$\mathrm{diag}(\mathbf{Q}) = \mathbf{I}_{10}$ and $\mathbf{R} = \mathbf{I}_8$.

\subsection{Learning Noise Models}

\begin{table*}
\centering
\caption{Results for planar pushing:
Translation (tr) and rotation (rot) error and negative log likelihood for the
DFs with different noise models (mean and standard error).
The hand-tuned DFs use fixed noise models whereas for the other variants, 
the noise models are trained end-to-end through the DFs. $\mathbf{R}_c$ indicates
a constant observation noise model and $\mathbf{R}_h$ a heteroscedastic one 
(same for $\mathbf{Q}$). The best result per DF and metric is highlighted in bold.
}\label{app:tab-pushing-noise}
\footnotesize
{
\begin{tabularx}{0.75\linewidth}{c l Y Y Y Y Y} 
\toprule
& & \mbox{Hand-tuned} $\mathbf{R}_c\mathbf{Q}_c$ &  $\mathbf{R}_c\mathbf{Q}_c$ &  $\mathbf{R}_h\mathbf{Q}_c$ &  $\mathbf{R}_c\mathbf{Q}_h$ & $\mathbf{R}_h\mathbf{Q}_h$ \\
\midrule
\multirow{5}*{\rotatebox[origin=c]{90}{tr [mm]}}  
& dEKF      & 6.22  & 4.45 & 4.61 & 4.44 & $\mathbf{4.38}$ \\
& dUKF      & 4.87  & 4.44 & 5.25 & $\mathbf{4.43}$ & 4.45 \\
& dMCUKF    & 4.73  & 4.42 & 4.8  & 4.39 & $\mathbf{4.35}$ \\
& dPF-M     & 18.13 & 5.07 & 4.92 & 5.32 & $\mathbf{4.64}$ \\
& dPF-G     & 17.95 & $\mathbf{5.48}$ & 35.57 & 210.45 & 10.92 \\
\midrule
\multirow{5}*{\rotatebox[origin=c]{90}{rot [$^{\circ}$]}}  
& dEKF      & 10.49            & 10.00 & $\mathbf{9.71}$  & 10.15 & 9.97 \\
& dUKF      & 9.87             & 9.91  & $\mathbf{9.73}$  & 10.05 & 10.00 \\
& dMCUKF    & $\mathbf{9.78}$  & 9.95  & 9.93             & 10.04 & 9.85 \\
& dPF-M     & 16.18            & 10.18 & $\mathbf{9.92}$  & 10.39 & 10.06 \\
& dPF-G     & 16.56 & 10.27 & 11.27    & 43.41 & $\mathbf{10.25}$ \\
\midrule
\multirow{5}*{\rotatebox[origin=c]{90}{NLL}}  
& dEKF      & 265.17 & 126.69 & 33.09  & 79.24  & $\mathbf{26.48}$ \\
& dUKF      & 378.08 & 84.12  & 33.06  & 81.55  & $\mathbf{27.61}$ \\
& dMCUKF    & 130.22 & 78.53  & 30.43  & 64.12  & $\mathbf{30.1}$ \\
& dPF-M     & 353.25 & 128.15 & 104.40 & 103.21 & $\mathbf{82.46}$ \\
& dPF-G     & $>$ 16m & 12,089.71 & 34.18 & 5,789.83 & $\mathbf{31.60}$ \\
\bottomrule
\end{tabularx}
}
\end{table*}

\begin{table*}
\centering
\caption{Results on pushing: Comparison between the DFs and LSTM. 
Process and sensor model are pretrained and get finetuned end-to-end. The DFs learn
heteroscedastic noise models. Each experiment 
is repeated three times and we report mean and standard errors.
}\label{app:push-tab-benchmark}
\footnotesize
\begin{tabularx}{0.6\linewidth}{X Y Y Y Y} 
\toprule
& RMSE  & NLL  & tr [mm] & rot [$^{\circ}$]  \\
\midrule
dEKF      &  14.9$\pm$0.46 & 33.9$\pm$3.86  & $\mathbf{3.5\pm0.02}$  & 8.8$\pm$0.22 \\
dUKF      &  $\mathbf{13.7\pm0.15}$ & $\mathbf{31.1\pm1.90}$  & 3.7$\pm$0.06  & 8.8$\pm$0.14 \\
dMCUKF    &  13.8$\pm$0.10 & 34.1$\pm$3.57  & 3.7$\pm$0.06  & $\mathbf{8.8\pm0.06}$ \\
dPF-M     &  18.3$\pm$0.38 & 120.4$\pm$5.70 & 5.7$\pm$0.16  & 10.5$\pm$0.36 \\
dPF-G     &  23.2$\pm$3.60 & 35.8$\pm$1.86  & 6.9$\pm$1.43  & 11.9$\pm$0.67 \\
dPF-M-lrn &  29.0$\pm$0.73 & 486.0$\pm$3.27 & 12.0$\pm$0.78 & 18.9$\pm$0.04 \\
dPF-G-lrn &  29.2$\pm$0.67 & 40.8$\pm$0.82  & 10.9$\pm$0.27 & 19.9$\pm$0.52  \\
LSTM      &  27.36$\pm$0.2 & 35.4$\pm$0.24  & 8.8$\pm$0.17  & 19.0$\pm$0.001  \\
\bottomrule
\end{tabularx}
\end{table*}

\paragraph{Experiment}

The diagonals of the hand-tuned models are 
\begin{align*}
&\sqrt{\mathrm{diag}(\mathbf{Q})} = \\
&\hspace{2em} \left(0.23 \hspace{0.5em} 0.23 \hspace{0.5em}  0.37 \hspace{0.5em}  0.01 \hspace{0.5em} 
 0.01 \hspace{0.5em}  0.7 \hspace{0.5em} 0.7 \hspace{0.5em} 0.1 \hspace{0.5em} 0.1 \hspace{0.5em} 0.13\right)^T
\end{align*} and 
\begin{align*}
&\sqrt{\mathrm{diag}(\mathbf{R})} = \\
&\hspace{2em}\begin{pmatrix} 
3.0 & 2.5 & 8.8 & 3.3 & 1.0 & 0.1 & 0.1 & 0.3
\end{pmatrix}^T
\end{align*}

\paragraph{Results}

Table~\ref{app:tab-pushing-noise} extends the results from 
Table~\ref{tab-pushing-noise} with data for the dPF-G. In contrast to 
the other DF variants, learning complex noise models for the 
pushing task is not successful for the dPF-G. While the NLL can be 
further decreased when the noise models are heteroscedastic instead of
constant, this comes at the cost of a significantly decreased tracking
performance.

\subsection{Benchmarking}

Table~\ref{app:push-tab-benchmark} extends the results from 
Table~\ref{push-tab-benchmark} with data for the dPF-G and dPF-G-lrn. Note that
their difference in performance to the dPF-M variants is smaller here than for
the previous tasks because training on $L_{\mathrm{mix}}$ instead of $L_{\mathrm{NLL}}$
reduces the effect of how the belief is represented on the loss.

\end{document}